\documentclass{article}

\PassOptionsToPackage{numbers, compress}{natbib}

\usepackage[main, final]{neurips_2026}

\usepackage[utf8]{inputenc} 
\usepackage[T1]{fontenc}    
\usepackage{hyperref}       
\usepackage{url}            
\usepackage{booktabs}       
\usepackage{amsfonts}       
\usepackage{nicefrac}       
\usepackage{microtype}      
\usepackage{xcolor}         

\usepackage{amsmath}
\usepackage{amssymb}
\usepackage{mathtools}
\usepackage{amsthm}

\usepackage{booktabs}
\usepackage{microtype}
\usepackage{graphicx}
\usepackage{booktabs} 
\usepackage{makecell}
\usepackage{amsmath}
\usepackage{verbatim}
\usepackage{epstopdf}
\usepackage{multirow}
\usepackage{amsfonts}
\usepackage{soul}
\usepackage{natbib}
\usepackage{wrapfig}
\usepackage{graphicx}
\usepackage{subcaption}

\title{Efficient Collaborative LLM Fine-Tuning over Heterogeneous Mobile Devices via Many Backbones to One Side-Network Tuning}

\author{
\textbf{
Xingke Yang\textsuperscript{1,*},
Liang Li\textsuperscript{2,*},
Sicong Li\textsuperscript{1},
Liwei Guan\textsuperscript{3},
Hao Wang\textsuperscript{5},
Xiaoqi Qin\textsuperscript{3},
Jiang Liu\textsuperscript{4}
}
\\
\textbf{
Xin Fu\textsuperscript{1},
Miao Pan\textsuperscript{1}
}
\\[0.5em]
$^{1}$University of Houston, USA \quad
$^{2}$Pengcheng Laboratory, China \quad
\\
$^{3}$Beijing University of Posts and Telecommunications, China
\\
$^{4}$Waseda University, Japan \quad
$^{5}$Stevens Institute of Technology, USA\quad
\\[0.35em]
\texttt{xyang56@cougarnet.uh.edu}
\quad
\texttt{mpan2@central.uh.edu}
}

\begin{document}

\maketitle
\begingroup
\renewcommand{\thefootnote}{}
\footnotetext{$^{*}$Equal Contribution}
\endgroup

\begin{abstract}

Collaboratively fine-tuning (FT) large language models (LLMs) over heterogeneous mobile devices fosters immense potential applications of personalized intelligence. However, such a vision faces critical system challenges. Existing federated LLM fine-tuning approaches remain limited by prohibitive on-device resource overhead, straggler-prone synchronous aggregation, and the assumption of a unified pretrained backbone across heterogeneous devices. In this paper, we propose EC-MobiLLM, a novel design for efficient collaborative LLM fine-tuning across heterogeneous mobile devices and heterogeneous backbones. EC-MobiLLM adopts a server-assisted side-tuning paradigm to minimize on-device overhead and pioneers non-blocking asynchronous collaboration to accelerate training. Moreover, EC-MobiLLM introduces adaptive feature alignment to support collaboration across heterogeneous backbones, aligning with practical needs in real-world deployments. Extensive experimental results demonstrate that EC-MobiLLM can maintain robust fine-tuning performance while achieving extremely low on-device memory, with at least 95.2\% reduction in computation overhead, 93.2\% reduction in communication costs and $5.1\times$ faster convergence compared to existing methods, validating its efficacy for practical LLM adaptation over heterogeneous mobile devices.

\end{abstract}

\section{Introduction}\label{sec:introduction}

\textcolor{black}{Fine-tuning (FT) large language models (LLMs) for domain-specific tasks unlocks significant potential for novel applications, driving growing demand for personalized intelligence~\citep{ren2024representation,ye2024cross,brown2020language}. 
In practice, the limited scale and diversity of data on an individual mobile device is often insufficient to support robust fine-tuning, motivating collaborative LLM fine-tuning across devices to improve model performance. Early works~\citep{zhang2023fedpetuning,tian2022fedbert} explored collaborative LLM FT by combining parameter-efficient fine-tuning (PEFT) methods (e.g., Adapters~\citep{houlsby2019adapter} and LoRA~\citep{hu2022lora}) with the traditional federated learning (FL)~\citep{mcmahan2017communication} paradigm. These approaches follow the standard FL loop (local training $\rightarrow$ upload $\rightarrow$ server aggregation $\rightarrow$ model distribution) and enable cross-device collaboration by exchanging only lightweight trainable modules.}

\textcolor{black}{However, local training in these methods faces critical bottlenecks due to the limited computational power and memory capacity of mobile devices. For example, tuning a model with 1.3B-parameters with LoRA typically requires more than 14.5 GB of GPU memory, which exceeds the 4–12 GB available on most mobile devices~\citep{li2025mobillm}. To further alleviate on-device burden, split federated learning (SFL) approaches further reduce device load by offloading deeper layers to the server~\citep{lin2024splitlora,chen2025memory}. However, its frequent exchange of activations and gradients further introduces substantial communication overhead. Moreover, the inherently synchronous aggregation protocol of FL and SFL requires strict sequential scheduling, leading to accumulated latency and dramatically prolonged convergence time—an issue amplified by the sheer size of modern LLMs.}

\textcolor{black}{Furthermore, in real-world deployments, participating mobile devices are often highly heterogeneous in computational power, memory capacity, and network bandwidth. To alleviate the resulting straggler effect, some studies~\citep{bai2024federated, cho2024heterogeneous} explore heterogeneous assignment of tuning modules, where devices are equipped with PEFT modules of different sizes (e.g., client-specific LoRA ranks) to adjust on-device computation, memory, and communication costs. However, these approaches enforce a unified pretrained backbone for all devices. Under this constraint, only adjusting LoRA ranks provides limited differentiation in on-device cost and fine-tuning performance. As a result, constraining high-resource devices to the same base model as low-resource ones wastes their capabilities. In practice, such devices should be able to deploy larger backbones to achieve stronger representations and a higher performance ceiling.}

To tackle the above issues, we propose EC-MobiLLM, a novel and efficient framework for collaborative LLM fine-tuning over heterogeneous mobile devices, built on the server-assisted side-tuning principle. Specifically, we decouple resource-intensive gradient computation from mobile devices by hosting all trainable parameters within a shared side-network on the server, while each mobile device retains only its own resource-appropriate frozen backbone LLM. During collaborative fine-tuning, each mobile device executes forward propagation computations on local data using its frozen backbone and uploads only selected intermediate activations to the server. The server performs asynchronous backpropagation using these activations, computing gradients and updating the shared side-network and lightweight projection matrices - without requiring global synchronization. In this way, EC-MobiLLM allows mobile devices to bypass costly on-device backpropagation and optimizer steps, drastically reducing client memory and computational load. By enabling server-side side-network training to proceed immediately upon receiving any mobile device’s activations, EC-MobiLLM eliminates the fundamental straggler bottleneck inherent in synchronous FL aggregation protocols. Crucially, EC-MobiLLM supports heterogeneous backbone sizes across devices while preserving cross-device collaborative LLM FT, enabling each device to benefit from the shared data diversity and achieve better fine-tuning performance.

Our salient contributions are summarized as follows:

 \textcolor{black}{(i) We propose EC-MobiLLM, a novel framework that pioneers mobile-friendly, asynchronous server-assisted side-tuning for LLM adaptation across distributed mobile data. Our design decouples computation by having devices perform only forward passes and upload activations, where the server asynchronously updates a unified side-network per-client activation arrival. This eliminates synchronization bottlenecks and removes all gradient computation from devices.}

\textcolor{black}{(ii) We design adaptive mechanisms that enable EC-MobiLLM to support heterogeneous mobile devices through cross-backbone alignment techniques. This ensures heterogeneous devices with different size models can collaboratively train a unified side-network that consolidates knowledge from all devices.}


  (iii) We implement and evaluate EC-MobiLLM across diverse mobile platforms (NVIDIA Jetson TX2, Xavier NX, and AGX Xavier) and model scales (sub-billion to billion parameters). Experiments across multiple tasks and system settings demonstrate that EC-MobiLLM achieves extremely low on-device memory usage, with at least 95.2\% reduction in computation overhead, 93.2\% reduction in communication costs and $5.1\times$ faster convergence compared to existing methods. It also delivers state-of-the-art and highly robust LLM fine-tuning performance.

\section{Background and Motivations}\label{sec:background}

\noindent\textbf{Inefficiencies of Federated LLM FT.} \textcolor{black}{The limited scale and diversity of data on individual mobile devices necessitate collaborative LLM fine-tuning across devices to enhance model performance. FL has emerged as a dominant paradigm for this purpose, where devices perform local training and upload parameter updates to a central server for aggregation. However, even with PEFT techniques such as Adapters~\citep{houlsby2019adapter}, LoRA~\citep{hu2022lora}, and BitFit~\citep{zaken2021bitfit}, locally fine-tuning LLMs under the FL paradigm remains challenging due to the substantial computation and memory demands of modern LLMs. To further alleviate on-device memory burden, split federated learning (SFL) approaches further reduce device load by offloading deeper layers to the server~\citep{lin2024splitlora,chen2025memory,gupta2018distributed} but introduces communication and coordination bottlenecks. Specifically, devices sequentially exchange activations/gradients with the server during forward/backward passes, while the device-side sub-models require periodic weight aggregations across devices. Alternatively, forward-only methods like FwdLLM~\citep{xu2024fwdllm} eliminate backpropagation by estimating gradients through parameter perturbations. While reducing activation memory to inference levels, this approach requires multiple forward passes per update - increasing device computational overhead. In addition, sequential execution of clients and servers in FL also hinders fast convergence in collaborative LLM fine-tuning. Collectively, existing federated LLM fine-tuning approaches exhibit critical gaps and cannot simultaneously optimize on-device memory usage, computational overhead, communication cost, and fine-tuning performance.}

\begin{wraptable}{r}{0.5\linewidth}
\centering
\small
\caption{Fine-tuning accuracy and memory footprint under different LoRA ranks and backbones (batch size: 16; sequence length: 256; Task: MNLI).}
\label{tab:lora_rank_compare}
\setlength{\tabcolsep}{2.5pt} 
\begin{tabular}{c|c|c|c|c|c}
\hline
\multicolumn{2}{c|}{Pre-trained Model} &
\multicolumn{3}{c|}{OPT-350M} &
OPT-1.3B \\
\hline
\multicolumn{2}{c|}{LoRA Rank (r)} &
r = 16 & r = 64 & r = 128 & r = 64 \\
\hline
\multicolumn{2}{c|}{Accuracy} &
 76.1 & 76.5 &76.8  &  81.1\\
\hline
\multicolumn{2}{c|}{Memory(GB)} &
 6.65 & 6.70 & 6.79 & 14.49 \\
\hline
\end{tabular}
\end{wraptable}


\noindent\textbf{Heterogeneity Challenges for collaborative LLM FT.}  \textcolor{black}{The significant variation in computational power, memory capacity, and network bandwidth across mobile devices introduces fundamental limitations to traditional synchronous federated learning protocols, which require the central server to wait for model updates from all participating devices before every-round global aggregation. This synchronization barrier creates unavoidable delays caused by slow devices (stragglers), forcing faster devices to remain idle during waiting periods~\citep{su2024dafl}. Moreover, memory heterogeneity also leads to significant resource waste. Existing heterogeneous collaborative LLM FT approaches such as FlexLoRA~\citep{bai2024federated}, FLoRA~\citep{wang2024flora}, and HetLoRA~\citep{cho2024heterogeneous} largely follow the standard FL+LoRA framework. They assume the same pretrained backbone across clients, configure client-specific LoRA ranks to match individual computing resource budgets, and design aggregation schemes to effectively combine LoRA updates with heterogeneous ranks. However, as shown in Table 1, the difference in training memory usage across various LoRA ranks is not significant. In practice, capable devices naturally prefer to load larger, more powerful models for achieving better performance. In summary, existing FL-based collaborative LLM fine-tuning approaches still face a significant applicability gap in real-world deployments. This calls for a breakthrough collaborative LLM fine-tuning framework that better accommodates heterogeneous devices and supports flexible asynchronous collaboration.}

\begin{wrapfigure}{rt}{0.5\linewidth}
    \centering
    \includegraphics[width=\linewidth]{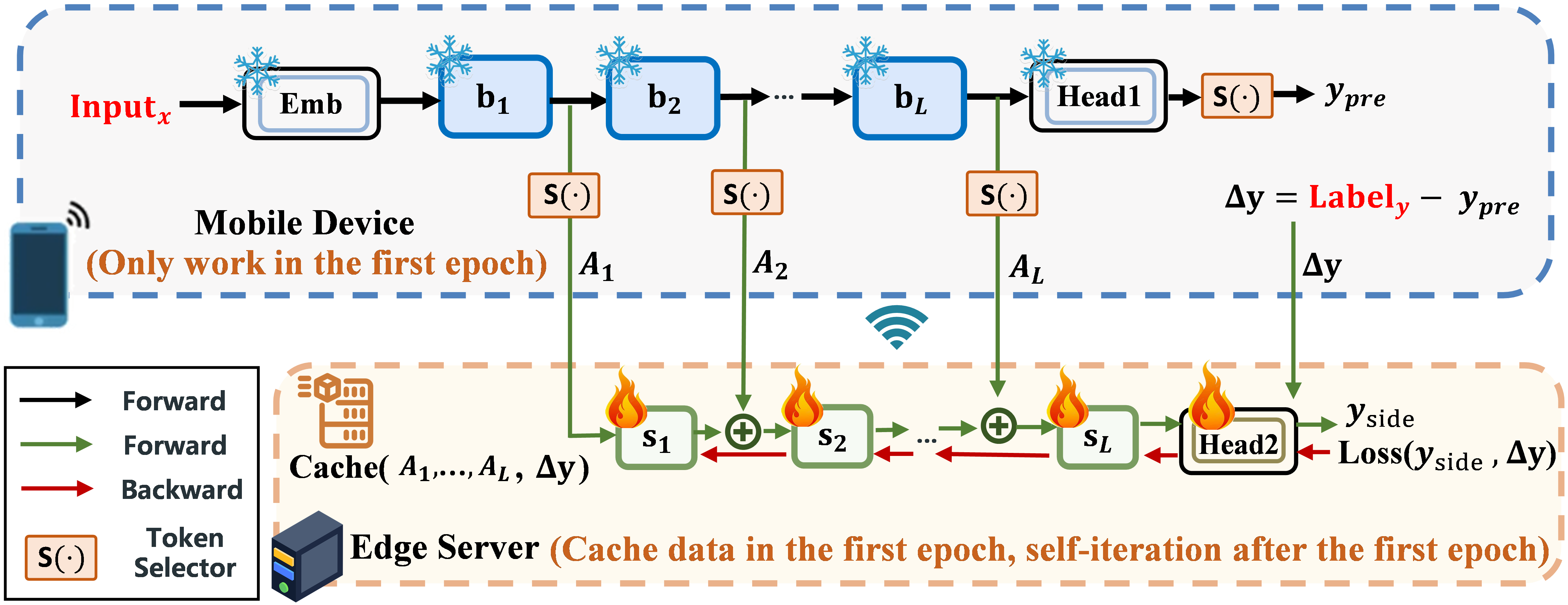}
    \caption{An overview of the PAE MobiLLM.}
    \label{fig:PAE}
\end{wrapfigure}

\noindent\textbf{Efficient On-device LLM Fine-Tuning.}  \textcolor{black}{Enabling LLM fine-tuning directly on mobile devices requires innovative architectures that minimize memory and computational load. Server-assisted side-tuning, pioneered by MobiLLM~\citep{li2025mobillm}, addresses this by decoupling trainable side-networks from frozen backbones and offloading all the gradient computation to the server. Different from split learning’s strictly sequential pipeline, MobiLLM allows the device and the server to compute in parallel. Advancing this approach, PAE MobiLLM~\citep{yang2025pae} (as illustrated in Fig.~\ref{fig:PAE}) introduces key optimizations: Mobile devices perform only a single forward pass through frozen backbones, compute the output deviation $\Delta y= \text{Label}_y- y_{\text{pre}}$ as the training target for the server-side side network, and upload token-selected intermediate activations $(\mathbf{A}_1, \dots, \mathbf{A}_L)$ together with $\Delta y$. The server then trains the side-network exclusively using these sparse activation-deviation pairs (\(\mathbf{A}_1, \cdots, \mathbf{A}_L, \Delta \mathbf{y} \)). A server-cached replay mechanism further reduces device overhead by limiting local data processing to the first epoch, with subsequent iterations handled server-side. This achieves an efficient balance between device resource consumption, communication costs, and training speed for on-device LLM fine-tuning. While highly effective in single-device settings, scaling to multi-device collaboration introduces fundamental challenges, including heterogeneity in both device resources and backbone models. We bridge this gap by extending the side-tuning paradigm to multi-device collaborative LLM fine-tuning with novel architectural innovations.}

\section{EC-MobiLLM Design}\label{sec:EC-MobiLLM Design}

\begin{figure*}
    \centering
    \includegraphics[width=0.99\textwidth]{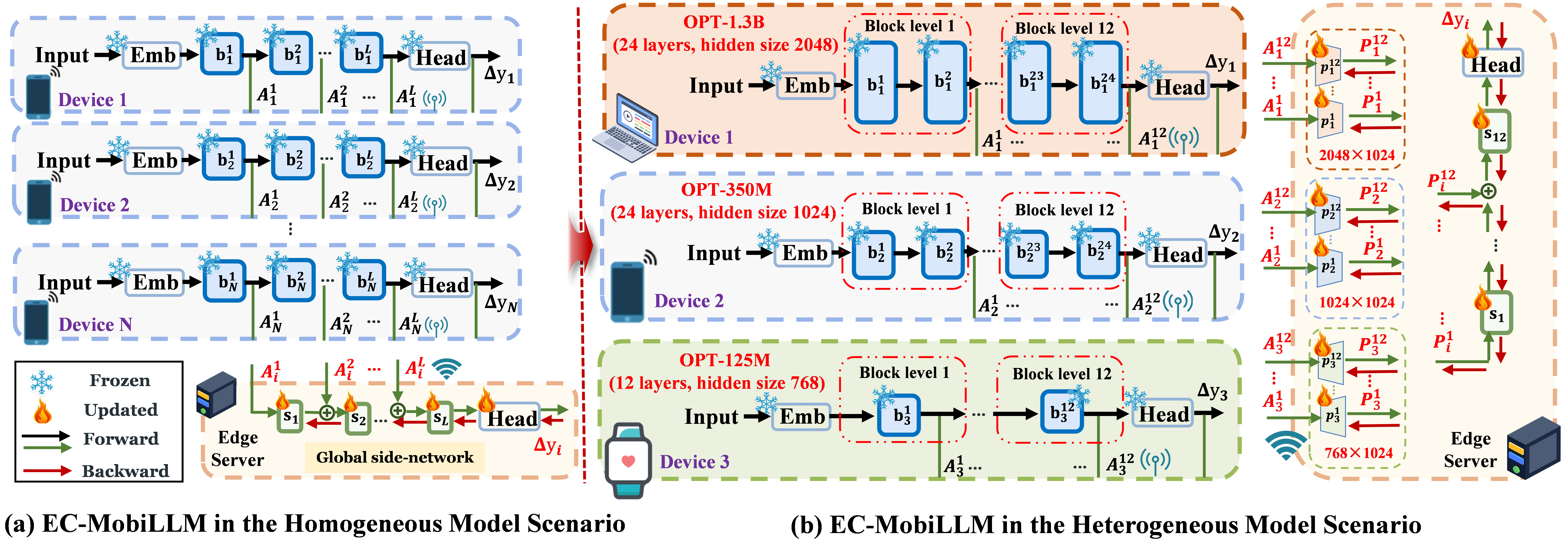}
    \caption{An overview of the EC-MobiLLM. }
    \label{fig:overview}
    \vspace{-1.1em}
\end{figure*}

\textcolor{black}{EC-MobiLLM is a collaborative LLM fine-tuning framework designed for heterogeneous mobile devices, supporting collaborative training with heterogeneous backbone models. Fig.~\ref{fig:overview} presents an overview of the EC-MobiLLM system. EC-MobiLLM provides a practical and efficient collaboration framework through inference-level on-device overhead, non-blocking asynchronous collaboration, and device-adaptive backbone selection, making it fundamentally different from conventional FL-based collaborative LLM fine-tuning methods. Specifically, each device retains only its own resource-appropriate frozen backbone LLM and performs a single forward pass using local data, without requiring on-device backpropagation. The device can then upload selected sparse intermediate activations and training target to the server at any time, without requiring synchronization between devices. The server trains a unified side network model using activation values uploaded from multiple devices, without requiring special parameter aggregation. Furthermore, the server can cache and reuse received activation values to independently and rapidly complete iterative training of the side network, without requiring continuous device-server data exchange.}

\textcolor{black}{Note that the concerns about data privacy arising from the transmission of intermediate activations in EC-MobiLLM are aligned with the definitions adopted in split learning related architectures~\citep{tian2022fedbert} and Google’s federated learning (FL) work~\citep{mcmahan2017communication}. Similar to these approaches, EC-MobiLLM is naturally compatible with existing advanced privacy-preserving techniques, such as differential privacy (DP) mechanisms~\citep{dwork2006differential} (e.g., adding DP noise to activations, gradients, outputs, or objective functions) and secure multiparty computation~\citep{du2001secure}. Further discussion is provided in Appendix~\ref{app:Privacy}.}

\subsection{Non-blocking asynchronous collaboration}

\textcolor{black}{Before collaborative training starts, devices communicate their backbone model specifications to the server, enabling it to initialize the side-network parameter structure and identify which intermediate activation layers each device should upload (Sec~\ref{sec:Heterogeneity-aware Cross-model Alignment}). During collaborative fine-tuning, devices and the server run in parallel, and no synchronization is required across devices:}

\noindent{\bf Mobile Devices.}
\textcolor{black}{Each mobile device performs forward propagation through its frozen backbone using mini-batches sampled from its local dataset. For each mini-batch, device \(i\) transmits selected intermediate activations (\(\mathbf{A}_i^1, \dots, \mathbf{A}_i^L \)) from the backbone layers, along with the prediction residual $\Delta y_i$  (defined as the difference between ground-truth label and backbone output) to the server as the side network training signal, consistent with PAE MobiLLM's design. Here, the selected activations mean that only the activations at token positions involved in loss computation are uploaded. For instance, for a classification task, this reduces the per-batch communication from 33.4 MB (full activations) to 0.78 MB~\citep{yang2025pae}. The mobile device continuously samples mini-batches from its local training set for a full epoch, performing local computation while simultaneously uploading activations. Since the backbone model on the mobile device remains frozen and is always kept in evaluation mode, stochastic layers such as dropout are disabled. Therefore, the intermediate activations produced for the same input samples stay unchanged across training iterations. As a result, after the device finishes one full pass over its local dataset, it does not need to perform redundant computations. Instead, the server can reuse the cached data for iterative computations.}


\noindent{\bf Server Training.}
\textcolor{black}{Upon receiving these activation-deviation pairs \((\mathbf{A}_i^1, \cdots, \mathbf{A}_i^L, \Delta \mathbf{y}_i)\) from any device, the server trains the side-network immediately, i.e., processing each device's contribution upon arrival without global synchronization. For each incoming sample, the server first checks the hidden size of the uploaded activations, applies the corresponding projection modules \(p\) in the forward pass, and then runs the shared side-network forward pass. It computes the loss between the side-network output and the deviation \(\Delta y\), and backpropagates along the same path. In other words, activations from different hidden-size backbones jointly train the shared side-network while separately updating their size-specific projection modules. After each update, the server caches received activations to construct an activation repository for iterative side-network training. During idle periods, the server trains continuously on cached samples to maximize computational efficiency. This design ensures uninterrupted training despite slow devices—eliminating straggler bottlenecks while maintaining full utilization of server resources. After all devices complete uploading, the server performs efficient standalone tuning using the comprehensive cached dataset. }

\textcolor{black}{After collaborative fine-tuning, devices download their specific projection modules alongside the shared side-network for local inference. These downloaded modules are lightweight, typically amounting to only about 1\%–2\% of the pretrained backbone parameters~\citep{li2025mobillm, yang2025pae}, and introduces only minor additional inference overhead on mobile devices.}


\subsection{\textcolor{black}{Heterogeneity-aware Cross-model Alignment}}\label{sec:Heterogeneity-aware Cross-model Alignment}

\textcolor{black}{Beyond computational heterogeneity through non-blocking device-server collaboration, EC-MobiLLM fundamentally resolves memory-driven model capacity divergence across mobile devices where deployable backbone sizes are dictated by each device's memory constraints. Specifically, EC-MobiLLM allows each device to load a pre-trained backbone model scaled to its hardware capacity. As illustrated in Fig. ~\ref{fig:overview} (b), devices may employ different-sized models (e.g., OPT-125M, OPT-350M, OPT-1.3B~\citep{zhang2022opt}), resulting in backbone architectures with varying numbers of transformer layers and hidden dimensions. This architectural diversity introduces two key challenges: (1) server-side adapters must accommodate diverse backbone structures, and (2) uploaded activations exhibit inconsistent shapes across devices. To resolve these issues, we introduce two structural alignment mechanisms—layer-wise activation sampling and hidden dimension scaling—that unify activation patterns across heterogeneous devices for collaborative training of the shared side-network on the server.}

\noindent{\bf Layer-Wise Activation Sampling.}
\textcolor{black}{Our layer-wise activation sampling mechanism addresses mismatches in transformer layer counts across backbone models. Inspired by empirical evidence in LST~\citep{sung2022lst} demonstrating that not all layers require dedicated adaptation modules, we propose selectively extracting activations from strategic layer positions.}

\textcolor{black}{As shown in Fig.~\ref{fig:overview}(b), when devices use backbones with differing layer counts (e.g., 12 vs. 24 layers), we partition all models into a fixed number of blocks (e.g., 12 blocks). Activations are then sampled exclusively from the final layer of each block. The server-side shared side-network is configured with an equal number of adapter modules, each processing one aligned activation block. Through comparative experiments of various strategies (Sec.~\ref{sec:Alignment}), we established optimal configuration guidelines: the block count is set to the minimum layer depth among participating models (e.g., 12 blocks for 12/24-layer models), while deeper models are partitioned at uniform intervals (e.g., sampling every other layer in a 24-layer backbone). This approach ensures layer-wise structural consistency while preserving representational capacity.}

\noindent{\bf Hidden Size Scaling.}
\textcolor{black}{Pre-trained models of different scales exhibit varying hidden sizes, preventing direct integration of their activations into a unified side-network adapter module. To resolve this, we introduce dedicated trainable lightweight bottleneck projection modules for each backbone model type, mapping activations to a consistent hidden size for shared side-network processing. The trainable bottleneck projection module can be formulated as $P = L_1 L_2$, where
$P \in \mathbb{R}^{d_{\mathrm{in}} \times d_{\mathrm{out}}}$,
$L_1 \in \mathbb{R}^{d_{\mathrm{in}} \times d_r}$ and
$L_2 \in \mathbb{R}^{d_r \times d_{\mathrm{out}}}$.}

\textcolor{black}{As shown in Fig.~\ref{fig:overview} (b), three distinct backbone models (hidden sizes: 2048, 1024, 768) require dimension standardization. We configure the shared side-network adapter modules with a target hidden size of 1024 and deploy projection modules (\(p_i^1, \dots, p_i^{12}\)) on the server for each model variant. Consider device 1 using OPT-1.3B (hidden size 2048): its activations (\(\mathbf{A}_1^1, \dots, \mathbf{A}_1^{12} \)) pass through corresponding projection modules (\(p_1^1, \dots, p_1^{12} \)) $\in \mathbb{R}^{2048*1024}$, producing transformed activations (\(\mathbf{P}_1^1, \dots, \mathbf{P}_1^{12} \)) with uniform 1024-dimensional features. These standardized activations then feed into the shared adapter modules (\(s_i^1, \dots, s_i^{12}\)), which enable collaborative training across heterogeneous models. Our empirical study (Sec.~\ref{sec:Alignment}) suggests selecting mid-range dimensions (e.g., 1024 for 768/1024/2048 scenarios) can optimize efficiency-accuracy balance in multi-device scenarios, while prioritizing larger dimensions can preserve representation capacity in scenarios involving only two model sizes.}


\section{Experimental Setup}\label{sec:Experimental Setup}


\noindent{\bf Implementation:} \textcolor{black}{The experimental testbed consists of a server with the NVIDIA A100 GPU and three types of heterogeneous client devices representing increasing computational capabilities for LLM processing: (1)  NVIDIA Jetson TX2 (8GB RAM, 1.3 TFLOPS), (2) NVIDIA Jetson Xavier NX (8GB RAM, 6 TFLOPS peak), and (3) NVIDIA Jetson AGX Xavier (16GB RAM, up to 10 TFLOPS peak). }


\noindent{\bf Models:}
\textcolor{black}{To systematically evaluate EC-MobiLLM's performance, we employ two representative pre-trained LLM architectures: i) decoder-based OPT series (OPT-1.3B, OPT-350M, OPT-125M), and ii) encoder-based RoBERTa series (RoBERTa-large(350M) and RoBERTa-base(125M)). This selection ensures architectural diversity while maintaining mobile compatibility. All models are initialized via HuggingFace Transformers~\citep{wolf2019huggingface}.}

\noindent{\bf Datasets:}
\textcolor{black}{We take the GLUE benchmark~\citep{wang2018glue} and DialogSum dataset~\citep{chen2021dialogsum} for the evaluation of NLP tasks, which are widely used in the fine-tuning research for LLM~\citep{zhang2023fedpetuning,sun2024improving, sung2022lst}. GLUE benchmark comprises eight tasks, including linguistic acceptability (CoLA ~\citep{warstadt2019neural}), sentiment analysis (SST-2~\citep{socher2013recursive}), similarity and paraphrase (MRPC~\citep{dolan2005automatically}, QQP~\citep{iyer2017first}, STS-B~\citep{cer2017semeval}), and natural language inference (MNLI~\citep{williams2017broad}, QNLI~\citep{rajpurkar2016squad}, RTE~\citep{bentivogli2009fifth}). \textcolor{black}{DialogSum includes summaries of real-world conversations on a diverse set of topics and scenarios to evaluate text-generation tasks. We use ROUGE scores (R1/R2/RL) as the accuracy metric.} Following FedPETuning~\citep{zhang2023fedpetuning}, we simulate non-IID data partitions using a Dirichlet distribution with concentration parameter $\alpha$, where lower $\alpha$ values induce higher label distribution shift. \textcolor{black}{(See Appendix ~\ref{app:dataset} for dataset details.)}}


\noindent{\bf Parameters:}
\textcolor{black}{Following FedPETuning~\citep{zhang2023fedpetuning}, we set the number of communication rounds to 100 and the number of local training epochs to 1 for all baselines under the FL paradigm. All configurations deploy 100 clients with balanced device-type distribution in heterogeneous settings. Experiments are conducted in a semi-emulation manner, where on-device resource consumption and training time are measured on real devices. For EC-MobiLLM and those centralized fine-tuning baselines, the number of training epochs is set to 20. To ensure fair comparison, all experiments share the same configurations unless specified: FP16 precision, batch size 8, learning rate 5e-4, maximum sequence length 256, and 60 Mbps in-lab Wi-Fi transmission speed. Additionally, LoRA and EC-MobiLLM employ rank-64 low-rank trainable modules by default, while FwdLLM uses 300 global perturbations per iteration.}


\noindent{\bf Baselines for Performance Comparison:} We compare EC-MobiLLM with two groups of baselines:

\textbf{Paradigm baselines}: i)  \textbf{FedPETuning}~\citep{zhang2023fedpetuning} (hereafter \textbf{FL}) ii) \textbf{SplitLoRA}~\citep{lin2024splitlora} (hereafter \textbf{SFL}) iii) \textbf{FwdLLM}~\citep{xu2024fwdllm}, which represent federated learning, split federated learning, and BP-free perturbation training, respectively. Each paradigm baseline is evaluated with two representative PEFT methods: i) \textbf{LoRA}~\citep{hu2022lora}: Inserts trainable low-rank matrices into frozen backbone networks. ii) \textbf{BitFit}~\citep{zaken2021bitfit}: Fine-tunes exclusively bias terms while freezing other pre-trained weights.

\textbf{Heterogeneity-aware baselines}: \textbf{FlexLoRA}~\citep{bai2024federated} and \textbf{FLoRA}~\citep{wang2024flora}, which support heterogeneous LoRA ranks under a shared backbone while still following the federated learning paradigm for collaboration. 
\begin{table*}[t]
\centering
\caption{\textcolor{black}{Comparison across methods: i) Per-device resource efficiency (memory/computation/communication); ii) Fine-tuning accuracy (centralized training vs. distributed training); and iii) Time-to-Accuracy (TTA) under homogeneous (TX2/Xavier/AGX Xavier clusters) and heterogeneous (Mixed) device configurations. Task: RoBERTa-Base@MRPC. \textit{Note: EC-MobiLLM uses uniform backbone models across devices in mixed settings to ensure a fair paradigm-level comparison.}}}
\label{table:method_comparison}
\small
\setlength{\tabcolsep}{1.7pt}
\begin{tabular}{c|c|c|c|c|c|c|c|c|c}
\hline
\multirow{2}{*}[-1.5ex]{\makecell{Methods}} &
\multirow{2}{*}{\makecell{On-device\\memory\\(GB)}} &
\multirow{2}{*}{\makecell{On-device\\comp.\\(TFLOPs)}} &
\multirow{2}{*}{\makecell{On-device\\comm.\\(MB)}} &
\multicolumn{2}{c|}{\makecell[t]{FT Performance (Acc.)}} &
\multicolumn{4}{c}{\makecell[t]{TTA@86.5(Mins)}} \\
\cline{5-6} \cline{7-10}
& & & & Centralized & Distributed & \makecell{TX2\\(Hom.)} & \makecell{Xavier\\(Hom.)} & \makecell{AGX\\(Hom.)} & \makecell{Mixed\\(Heter.)} \\
\hline
FL-LoRA           & 1.18 & 395.7 & 31.1 & 91.5 & 86.9 & 54.2 & 41.6 &  35.7  & 49.3   \\ \hline
FL-BitFit         & 1.02 & 388.1 & 23.1 & 90.7 & 86.8 & 49.7 & 39.1 & 32.2 & 35.6   \\ \hline
SFL-LoRA          & 0.54 & 58.3 & 5437.5 & 91.5 & 87.8 & 782.1 & 763.2 & 741.3 &  777.3 \\ \hline
SFL-BitFit        & 0.49 & 56.5 & 5429.4 & 90.7 & 87.2 & 774.2 & 751.2 &  722.9 & 767.9 \\ \hline
FwdLLM-LoRA       & 0.42 & 407.4 & 22.8 & 91.5 & 87.1 & 41.2 & 28.9&  25.4&   40.3 \\ \hline
FwdLLM-BitFit     & \textbf{0.38} & 391.5 & 16.2 & 90.7 & 87.4 & 32.7 & 22.9 & 20.2& 30.3   \\ \hline
\textbf{EC-MobiLLM} & \textbf{0.38} & \textbf{2.7} & \textbf{1.1} & 89.6 & \textbf{88.1} & \textbf{5.7} & \textbf{6.0} & \textbf{5.1}  & \textbf{5.9}   \\
\hline
\end{tabular}
\end{table*}

\section{Evaluation Results and Analysis}\label{sec:Evaluation Results}

\subsection{Comparative Analysis with Federated FT Baselines}

\textcolor{black}{We conduct comprehensive experiments to validate EC-MobiLLM's advantages in on-device resource efficiency, training efficiency, and fine-tuning performance across homogeneous and heterogeneous device environments.} \textcolor{black}{(See Appendix ~\ref{app:calc} for details of result computation.)}


\noindent{\bf On-Device Resource Efficiency.}
\textcolor{black}{As detailed in Table~\ref{table:method_comparison}, EC-MobiLLM demonstrates superior resource efficiency across metrics in terms of on-device memory footprint, computation cost, and communication cost. Conventional FL-based PEFT paradigms incur substantial resource demands, while SFL approaches trade computation savings for significantly increased communication overhead (up to $175\times$). Similarly, FwdLLM trades memory savings for higher computation load. In contrast, EC-MobiLLM maintains an optimal balance, where devices perform only a single forward pass, reducing memory consumption to inference levels (e.g., $2.68\times$ reduction for RoBERTa-Base). Besides, it achieves at least 95.2\% lower computation and 93.2\% less communication by eliminating on-device backpropagation and leveraging server-side activation caching.}

\begin{figure*}[t]
\centering
\includegraphics[width=0.95\textwidth]{Pic/combined_plot_ICML.png} 
\caption{\textcolor{black}{Convergence performance on various models and tasks under heterogeneous-device settings.}} 
\label{Fig:Conv}
\vspace{-8pt} 
\end{figure*}



\begin{table*}[t]
\centering
\footnotesize
\caption{\textcolor{black}{FT Performance under data heterogeneity: IID vs. non-IID ($\alpha=\{0.1,1.0,10.0\}$).}}
\label{table:sst2_hetero}
\setlength{\tabcolsep}{3pt}
\begin{tabular}{c|c|c|c|c|c|c|c}
\hline
 & \multicolumn{4}{c|}{RoBERTa-Base@MRPC (Acc.)} & \multicolumn{3}{c}{OPT-1.3B@DialogSum (R1/R2/RL)} \\ \cline{2-8}
\multirow{2}{*}[+2ex]{Methods} &
IID &
\begin{tabular}[c]{@{}c@{}}non-IID\\($\alpha$= 10.0)\end{tabular} &
\begin{tabular}[c]{@{}c@{}}non-IID\\($\alpha$= 1.0)\end{tabular} &
\begin{tabular}[c]{@{}c@{}}non-IID\\($\alpha$= 0.1)\end{tabular} &
IID &
\begin{tabular}[c]{@{}c@{}}non-IID\\($\alpha$= 10.0)\end{tabular} &
\begin{tabular}[c]{@{}c@{}}non-IID\\($\alpha$= 0.1)\end{tabular} \\
\hline
FL-LoRA        & 86.9 & 85.1 & 84.5 & 82.1 & 19.2 / 6.1 / 15.1 & 17.9 / 5.2 / 14.3 &  16.1 / 4.6 / 12.7 \\ \hline
FL-BitFit      & 86.8 & 84.9 & 84.6 & 81.5 & 19.0 / 6.2 / 14.9  & 17.7 / 5.1 / 14.2 & 16.4 / 4.2 / 12.9 \\ \hline
SFL-LoRA       & 87.8 & 87.1 & 86.8 & 84.3 & 19.9 / 6.5 / 15.8  & 18.5 / 6.1 / 15.1 & 18.0 / 5.5 / 13.8 \\ \hline
SFL-BitFit     & 87.2 & 87.0 & 86.1 & 85.2 & 19.7 / 6.2 / 15.5  & 18.7 / 6.0 / 15.2 & 17.8 / 5.4 / 13.4 \\ \hline
FwdLLM-LoRA    & 87.1 & 86.7 & 86.5 & 84.1 & 19.3 / 6.3 / 15.7  & 18.2 / 6.2 / 15.1 & 17.8 / 5.2 / 13.7 \\ \hline
FwdLLM-BitFit  & 87.4 & 87.1 & 86.6 & 84.7 & 19.6 / 5.9 / 15.1  & 18.6 / 6.1 / 14.9 & 17.3 / 5.1 / 14.1 \\ \hline
\textbf{EC-MobiLLM} & \textbf{88.1} & \textbf{87.8} & \textbf{87.7} & \textbf{87.3} & \textbf{21.0 / 7.7 / 17.2}  & \textbf{20.2 / 7.4 / 16.0} & \textbf{20.0 / 7.3 / 16.3} \\
\hline
\end{tabular}
\vspace{-8pt} 
\end{table*}

\noindent{\bf Training Efficiency.}
\textcolor{black}{We evaluate EC-MobiLLM's training efficiency by measuring time-to-accuracy across diverse configurations, including two model architectures (OPT, RoBERTa) and two tasks (MRPC, QNLI), as shown in Fig.~\ref{Fig:Conv}. To ensure fair comparison of different collaboration paradigms, all heterogeneous devices use identical backbone models across EC-MobiLLM and other paradigm baselines. The results show that EC-MobiLLM achieves at least a $5.1\times$ speedup across all tasks. This acceleration stems from our full-pipeline efficient design: during the initial phase, clients perform only one forward propagation per data sample, avoiding both on-device backpropagation in conventional FL and multi-pass perturbations in FwdLLM. After receiving activations from all devices, the server performs iterative training on the shared side-network independently, eliminating parameter synchronization with devices inherent in FL paradigms.}

\textcolor{black}{We further evaluate training efficiency across different client device setups, as shown in the right portion of Table~\ref{table:method_comparison}, which validates EC-MobiLLM's superior straggler resilience. Under the heterogeneous-device setting, other collaboration paradigms suffer severe slowdowns due to synchronous waiting periods in federated aggregation protocols, resulting in training times approaching the all-TX2 (lowest-capacity devices) configuration. In contrast, EC-MobiLLM benefits from extremely lightweight on-device computation and non-blocking parallel device-server collaboration. As a result, computational speed variations across devices are effectively masked, yielding consistent training times across diverse client setups.}

\begin{figure}[t]
\centering
\begin{subfigure}[t]{0.24\textwidth}
  \includegraphics[width=\linewidth]{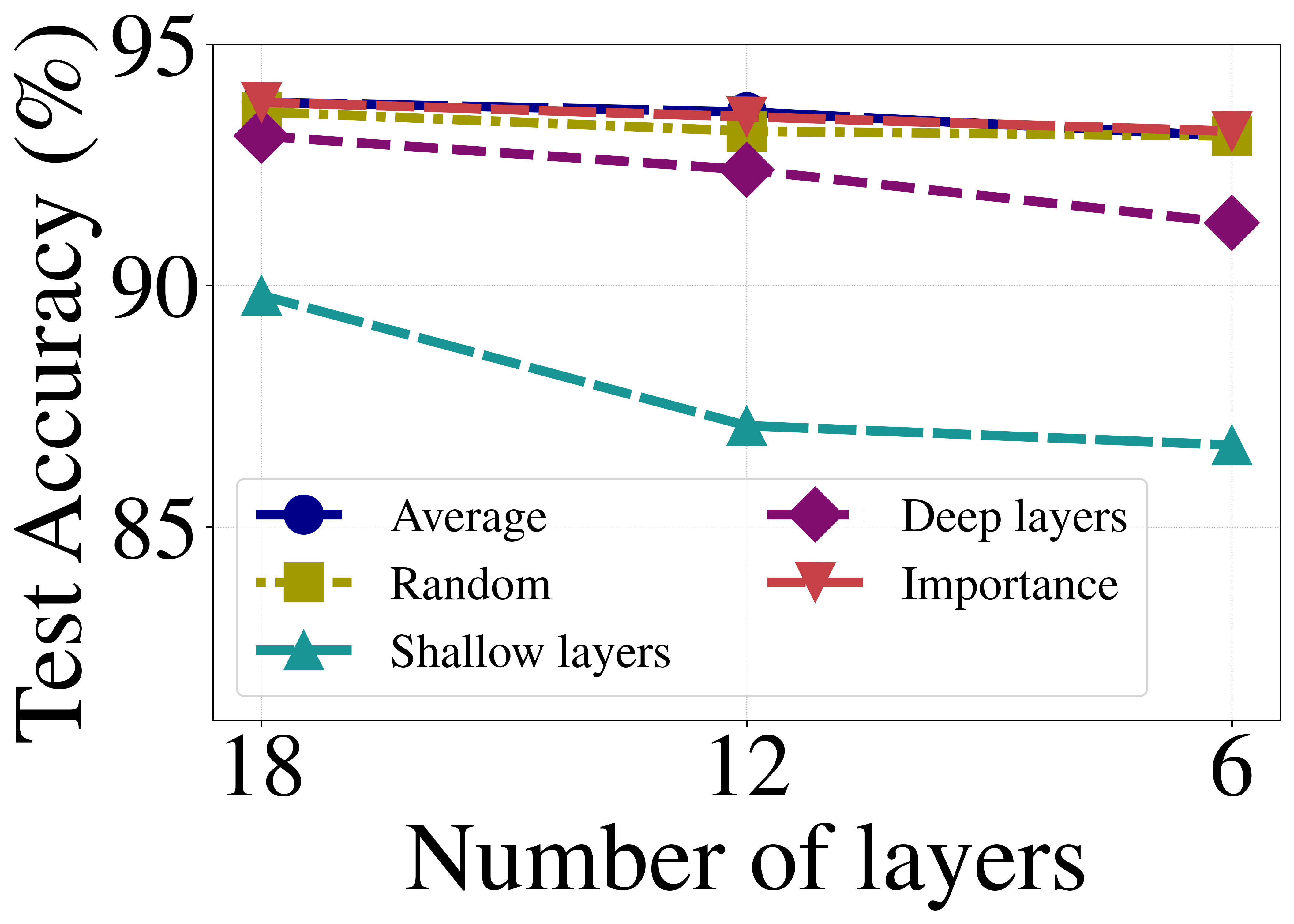}
  \caption*{\small (a) OPT-1.3B}
\end{subfigure}\hfill
\begin{subfigure}[t]{0.24\textwidth}
  \includegraphics[width=\linewidth]{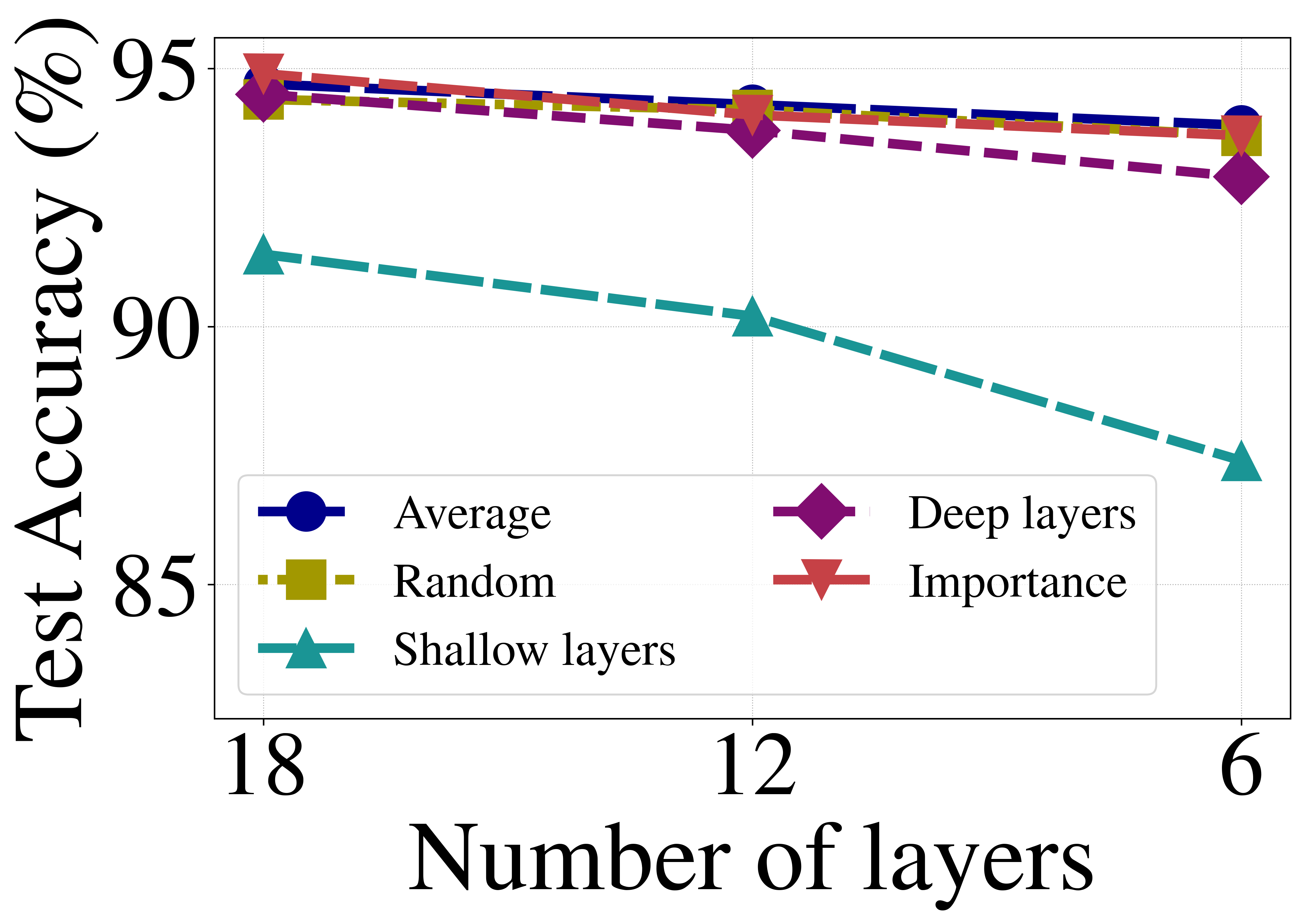}
  \caption*{\small (b) RoBERTa-large}
\end{subfigure}\hfill
\begin{subfigure}[t]{0.24\textwidth}
  \includegraphics[width=\linewidth]{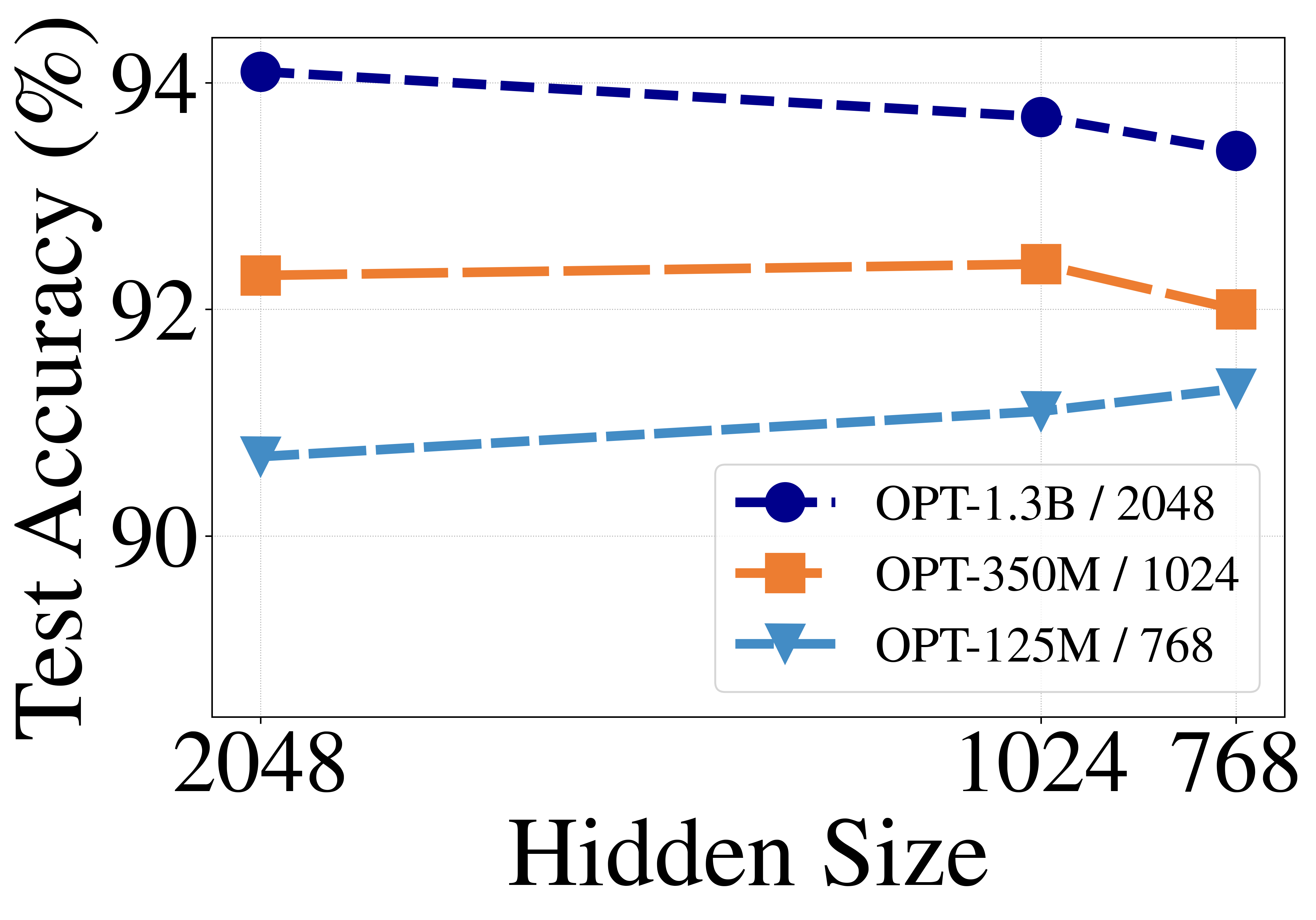}
  \caption*{\small (c) OPT Models}
\end{subfigure}\hfill
\begin{subfigure}[t]{0.24\textwidth}
  \includegraphics[width=\linewidth]{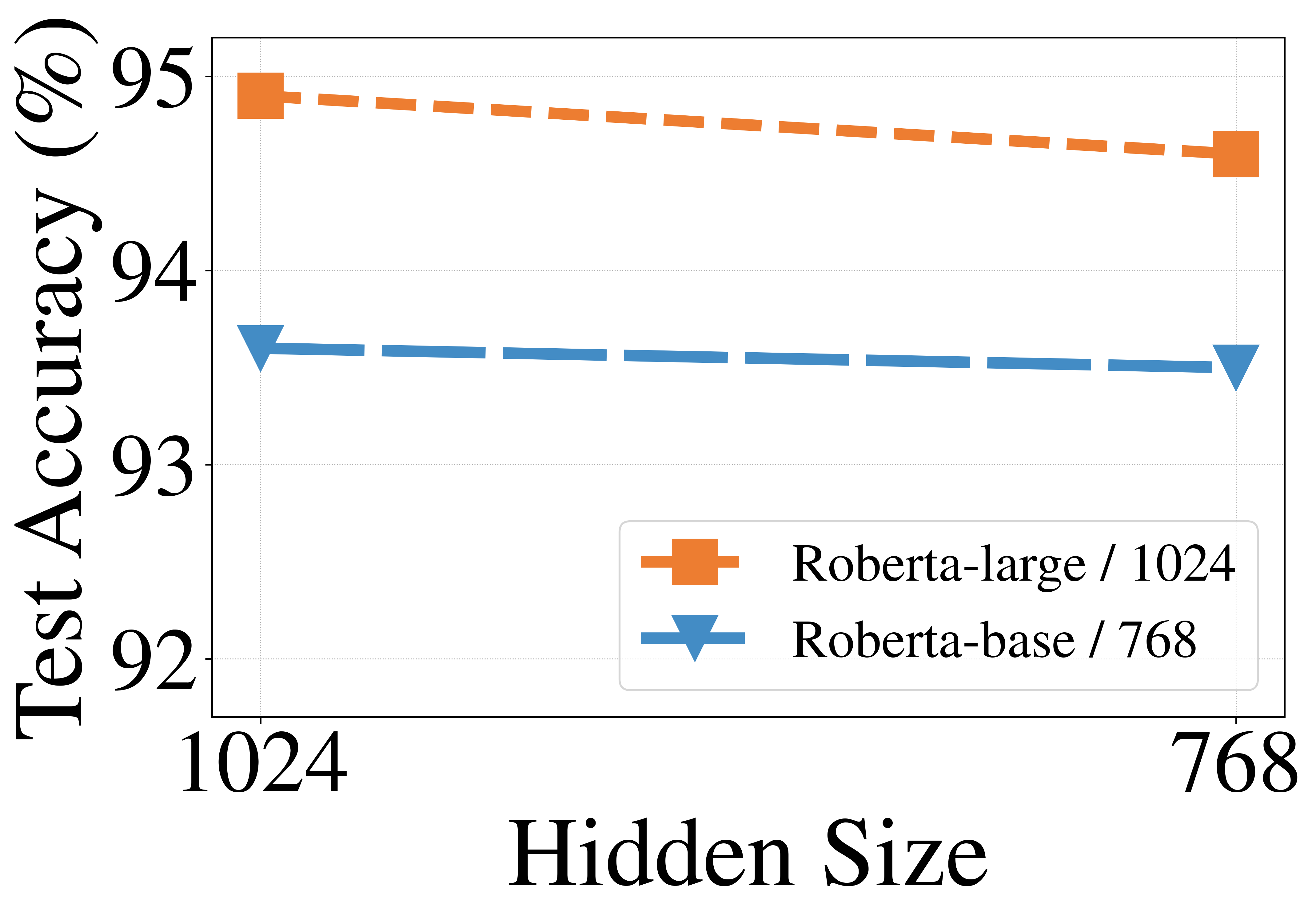}
  \caption*{\small (d) RoBERTa Models}
\end{subfigure}
\caption{\textcolor{black}{Comparative accuracy across sampling methods (a-b) and hidden-size scaling (c-d) (Task: SST-2).}}
\label{Fig:four-in-row}
\end{figure}

\begin{table*}[t]
\centering
\footnotesize
\caption{\textcolor{black}{Performance comparison of heterogeneity-aware methods under heterogeneous devices. (\textit{Single}:  isolated side-network training on each single device's local data; \textit{Global}: collaborative training of a shared side-network. Task: SST-2.) }}
\label{table:device_perf}
\setlength{\tabcolsep}{2pt}
\begin{tabular}{c|c|c|c|c|c|c|c}
\hline
\multirow{2}{*}{\makecell{Method}} &
\multirow{2}{*}{\makecell{Devices}} &
\multirow{2}{*}{\makecell{Local Backbone }} &
\multirow{2}{*}{\makecell{LoRA\\Rank ($r$)}} &
\multirow{2}{*}{\makecell{On-device\\Memory (GB)}} &
\multirow{2}{*}{\makecell{Per-device Local\\Runtime(Mins)}} &
\multicolumn{2}{c}{FT Performance (Acc.)} \\
\cline{7-8}
& & & & & & \multicolumn{1}{c|}{Single} & \multicolumn{1}{c}{Global} \\
\hline

\multirow{3}{*}{\makecell{FlexLoRA}}
& AGX    & OPT-125M      & 128 & 2.51 & 12.7 & 88.5 & 89.9 \\ \cline{2-8}
& Xavier & OPT-125M      & 64 & 2.48 & 15.8 & 88.5 & 89.7 \\ \cline{2-8}
& TX2    & OPT-125M      & 16 & 2.46 & 17.9 & 88.4 & 89.4 \\
\Xhline{1.5pt}

\multirow{3}{*}{\makecell{FLoRA}}
& AGX    & OPT-125M      & 128 & 2.51 & 12.7 & 88.8 & 90.8 \\ \cline{2-8}
& Xavier & OPT-125M      & 64 & 2.48 & 15.8 & 88.6 & 90.7 \\ \cline{2-8}
& TX2    & OPT-125M      & 16 & 2.46 & 17.9 & 88.6 & 90.4 \\
\Xhline{1.5pt}

\multirow{3}{*}{\makecell{EC-MobiLLM}}
& AGX    & OPT-1.3B      & -- & 3.44 & 1.53 & 90.0 & 92.9 \\ \cline{2-8}
& Xavier & OPT-350M      & -- & 1.10 & 1.46 & 89.9 & 92.4 \\ \cline{2-8}
& TX2    & OPT-125M      & -- & 0.56 & 1.41 & 88.4 & 92.1 \\
\hline
\end{tabular}
\vspace{-4pt} 
\end{table*}

\noindent{\bf LLM FT Performance.}
\textcolor{black}{We evaluate EC-MobiLLM against federated LLM FT methods and their centralized PEFT counterparts (LoRA, BitFit, side-network tuning). As Table~\ref{table:method_comparison} shows, while LoRA and BitFit outperform side-network tuning in centralized settings, their accuracy significantly degrades under distributed deployment with distributed data. Even the best federated baseline (FwdLLM-BitFit) exhibits at least 3.3\% performance drop. In contrast, EC-MobiLLM maintains near-centralized performance with only a 1.5\% accuracy drop, outperforming all federated FT baselines by at least 0.3\%.}

\textcolor{black}{We further evaluate the impacts of data heterogeneity on EC-MobiLLM’s performance. To build local datasets, we use three Dirichlet distributions ($\alpha \in \{0.1,1.0,10.0\}$)  where a lower $\alpha$ indicates a higher non-IID level~\citep{zhang2023fedpetuning}. Table~\ref{table:sst2_hetero} demonstrates that while the performance of all methods degrades under data heterogeneity, EC-MobiLLM shows superior resilience. Taking the MRPC task as an example, at extreme heterogeneity ($\alpha=0.1$), EC-MobiLLM’s accuracy drop is 1.2\% smaller than SFL-BitFit, validating enhanced robustness to cross-client data distribution shifts.}

\subsection{\textcolor{black}{Cross-Model Alignment Performance}}
\label{sec:Alignment}

\textcolor{black}{We evaluate the efficacy of our layer-wise activation sampling and hidden size scaling methods for heterogeneous model adaptation through comparative experiments with alternative approaches.}

\noindent{\bf Layer Alignment Design.}
\textcolor{black}{To determine optimal block configurations for layer alignment in EC-MobiLLM, we perform comparative experiments using OPT-1.3B and RoBERTa-Large models. Specifically, we feed sampled layer-wise backbone activations to the side network and analyze how layer-wise activation sampling strategies affect performance, as shown in Fig.~\ref{Fig:four-in-row} (a-b). We compare five strategies: shallow-only, deep-only, average-interval, random, and importance-based selection (identified via layer-wise ablation). Results indicate: i) Performance improves with more backbone activation layers; ii) The average-interval strategy achieves accuracy comparable to computationally intensive importance-based approaches. These findings support EC-MobiLLM's configuration: set block count to the most lightweight LLM backbone's layer depth across devices, partitioning larger-sized LLMs at equal intervals to ensure structural alignment while maintaining performance.}

\noindent{\bf Hidden Size Alignment Design.}
\textcolor{black}{For cross-model hidden size scaling, determining a unified dimension for side-network adapters is critical. We experiment with various OPT and RoBERTa models, and evaluate performance under different side-network hidden sizes. As shown in Fig.~\ref{Fig:four-in-row} (c-d), peak performance is achieved when side-network hidden sizes match backbone sizes, while dimensional mismatches degrade accuracy. For example, OPT-125M (hidden size =768) performs worse when forced dimension to 2048, which demonstrates that larger dimensions aren't always beneficial. These findings yield practical EC-MobiLLM configuration guidelines: for multi-device scenarios, select median dimensions (e.g., 1024 for 768/1024/2048); for dual-model scenarios, prioritize larger dimensions to preserve representational capacity.}

\subsection{Validation of Heterogeneous Backbone Adaptation}
\textcolor{black}{To validate the efficacy and necessity of EC-MobiLLM's heterogeneous-backbone design, we conduct systematic experiments. In EC-MobiLLM, each device loads a backbone model tailored to its hardware capabilities. In contrast, existing heterogeneous FL baselines are constrained to a single shared backbone (determined by the weakest device) and can only adjust the LoRA rank.}

\noindent{\bf Device-Specific Workload Balancing.}
\textcolor{black}{As shown in the results of the on-device workload in Table~\ref{table:device_perf}, device-specific backbone assignment optimizes hardware resource utilization compared to uniform model deployment. For example, AGX Xavier and Xavier run OPT-1.3B and OPT-350M, respectively, with memory usage both around 30\% of their available capacities (12.4 GB and 4.6 GB). Computational loads similarly scale to device capacities, resulting in comparable execution times across heterogeneous devices. This helps ensure near-balanced contributions to side-network training and prevent representation drift toward data from those fast devices. In contrast, devices still exhibit pronounced cross-device imbalance in FlexLoRA and FLoRA.}




\noindent{\bf Cross-Capacity Collaboration Effects.}
\textcolor{black}{We investigate whether collaboration with lower-capacity devices in EC-MobiLLM compromises high-capacity device performance. To this end, we evaluate performance under two settings: i) training the side network using only each device’s local data (denoted by \textit{Single}), and ii) collaboratively training a shared side network across all devices (denoted by \textit{Global}). As shown in Table~\ref{table:device_perf}, \textit{Global} consistently outperforms \textit{Single} across different backbone sizes, indicating that EC-MobiLLM enables all devices to benefit from a robust shared side network trained collaboratively on diverse cross-device data.  In addition, compared to FlexLoRA and FLoRA, a high-resource device (e.g., AGX) can achieve significantly better performance for both itself and its collaborating devices when using a larger backbone model. This highlights the importance of supporting device-adaptive backbone selection in collaborative LLM fine-tuning.}


\section{Conclusion}


\textcolor{black}{This paper has presented EC-MobiLLM, an efficient framework for collaborative LLM fine-tuning across heterogeneous mobile devices. By pioneering an asynchronous server-assisted side-tuning paradigm, EC-MobiLLM decouples device responsibilities to forward-only propagation and activation uploading, while the server trains a shared side-network, which eliminates synchronization bottlenecks inherent in conventional FL-based fine-tuning approaches. Through layer-wise activation sampling and cross-architecture dimension alignment, EC-MobiLLM enables each device to load backbone models that match its hardware capacities while still maintaining the ability to collaboratively train a shared side network, ensuring robust support for device heterogeneity. Extensive experiments demonstrate EC-MobiLLM's efficacy and efficiency: achieving $2.68\times$ reduction in on-device memory usage, 95.2\% reduction in computational cost, 93.2\% lower communication overhead, and $5.1\times$ faster convergence compared to existing methods, while maintaining competitive accuracy. Additional discussions on server-side cache storage, network-speed sensitivity, and privacy considerations for transmitted activations are provided in Appendix~\ref{app:Discussion}. These results collectively establish EC-MobiLLM as a practical and deployment-ready solution for real-world collaborative LLM fine-tuning on distributed mobile datasets.} 

\newpage

\bibliography{ref}

\begin{thebibliography}{40}
\providecommand{\natexlab}[1]{#1}
\providecommand{\url}[1]{\texttt{#1}}
\expandafter\ifx\csname urlstyle\endcsname\relax
  \providecommand{\doi}[1]{doi: #1}\else
  \providecommand{\doi}{doi: \begingroup \urlstyle{rm}\Url}\fi

\bibitem[Bai et~al.(2024)Bai, Chen, Qian, Yao, and Li]{bai2024federated}
Jiamu Bai, Daoyuan Chen, Bingchen Qian, Liuyi Yao, and Yaliang Li.
\newblock Federated fine-tuning of large language models under heterogeneous tasks and client resources.
\newblock \emph{Advances in Neural Information Processing Systems}, 37:\penalty0 14457--14483, 2024.

\bibitem[Bentivogli et~al.(2009)Bentivogli, Clark, Dagan, and Giampiccolo]{bentivogli2009fifth}
Luisa Bentivogli, Peter Clark, Ido Dagan, and Danilo Giampiccolo.
\newblock The fifth pascal recognizing textual entailment challenge.
\newblock \emph{TAC}, 7\penalty0 (8):\penalty0 1, 2009.

\bibitem[Brown et~al.(2020)Brown, Mann, Ryder, Subbiah, Kaplan, Dhariwal, Neelakantan, Shyam, Sastry, Askell, et~al.]{brown2020language}
Tom Brown, Benjamin Mann, Nick Ryder, Melanie Subbiah, Jared~D Kaplan, Prafulla Dhariwal, Arvind Neelakantan, Pranav Shyam, Girish Sastry, Amanda Askell, et~al.
\newblock Language models are few-shot learners.
\newblock \emph{Advances in neural information processing systems}, 33:\penalty0 1877--1901, 2020.

\bibitem[Cer et~al.(2017)Cer, Diab, Agirre, Lopez-Gazpio, and Specia]{cer2017semeval}
Daniel Cer, Mona Diab, Eneko Agirre, Inigo Lopez-Gazpio, and Lucia Specia.
\newblock Semeval-2017 task 1: Semantic textual similarity-multilingual and cross-lingual focused evaluation.
\newblock \emph{arXiv preprint arXiv:1708.00055}, 2017.

\bibitem[Chen et~al.(2025)Chen, Li, Ji, and Wu]{chen2025memory}
Xiaopei Chen, Liang Li, Fei Ji, and Wen Wu.
\newblock Memory-efficient split federated learning for llm fine-tuning on heterogeneous mobile devices.
\newblock In \emph{IEEE INFOCOM 2025-IEEE Conference on Computer Communications Workshops (INFOCOM WKSHPS)}, pages 1--6. IEEE, 2025.

\bibitem[Chen et~al.(2021)Chen, Liu, Chen, and Zhang]{chen2021dialogsum}
Yulong Chen, Yang Liu, Liang Chen, and Yue Zhang.
\newblock Dialogsum: A real-life scenario dialogue summarization dataset.
\newblock \emph{arXiv preprint arXiv:2105.06762}, 2021.

\bibitem[Cho et~al.(2024)Cho, Liu, Xu, Fahrezi, and Joshi]{cho2024heterogeneous}
Yae~Jee Cho, Luyang Liu, Zheng Xu, Aldi Fahrezi, and Gauri Joshi.
\newblock Heterogeneous lora for federated fine-tuning of on-device foundation models.
\newblock In \emph{Proceedings of the 2024 conference on empirical methods in natural language processing}, pages 12903--12913, 2024.

\bibitem[Dolan and Brockett(2005)]{dolan2005automatically}
Bill Dolan and Chris Brockett.
\newblock Automatically constructing a corpus of sentential paraphrases.
\newblock In \emph{Third international workshop on paraphrasing (IWP2005)}, 2005.

\bibitem[Du and Atallah(2001)]{du2001secure}
Wenliang Du and Mikhail~J Atallah.
\newblock Secure multi-party computation problems and their applications: a review and open problems.
\newblock In \emph{Proceedings of the 2001 workshop on New security paradigms}, pages 13--22, 2001.

\bibitem[Dwork(2006)]{dwork2006differential}
Cynthia Dwork.
\newblock Differential privacy.
\newblock In \emph{International colloquium on automata, languages, and programming}, pages 1--12. Springer, 2006.

\bibitem[Geiping et~al.(2020)Geiping, Bauermeister, Dr{\"o}ge, and Moeller]{geiping2020inverting}
Jonas Geiping, Hartmut Bauermeister, Hannah Dr{\"o}ge, and Michael Moeller.
\newblock Inverting gradients-how easy is it to break privacy in federated learning?
\newblock \emph{Advances in neural information processing systems}, 33:\penalty0 16937--16947, 2020.

\bibitem[Gupta and Raskar(2018)]{gupta2018distributed}
Otkrist Gupta and Ramesh Raskar.
\newblock Distributed learning of deep neural network over multiple agents.
\newblock \emph{Journal of Network and Computer Applications}, 116:\penalty0 1--8, 2018.

\bibitem[Houlsby et~al.(2019)Houlsby, Giurgiu, Jastrzebski, Morrone, De~Laroussilhe, Gesmundo, Attariyan, and Gelly]{houlsby2019adapter}
Neil Houlsby, Andrei Giurgiu, Stanislaw Jastrzebski, Bruna Morrone, Quentin De~Laroussilhe, Andrea Gesmundo, Mona Attariyan, and Sylvain Gelly.
\newblock Parameter-efficient transfer learning for nlp.
\newblock In \emph{Proc. of 36th International Conference on Machine Learning (ICML)}, Long Beach, CA, June 2019.

\bibitem[Hu et~al.(2022)Hu, Shen, Wallis, Allen-Zhu, Li, Wang, Wang, Chen, et~al.]{hu2022lora}
Edward~J Hu, Yelong Shen, Phillip Wallis, Zeyuan Allen-Zhu, Yuanzhi Li, Shean Wang, Lu~Wang, Weizhu Chen, et~al.
\newblock Lora: Low-rank adaptation of large language models.
\newblock \emph{ICLR}, 1\penalty0 (2):\penalty0 3, 2022.

\bibitem[Iyer et~al.(2017)Iyer, Dandekar, Csernai, et~al.]{iyer2017first}
Shankar Iyer, Nikhil Dandekar, Korn{\'e}l Csernai, et~al.
\newblock First quora dataset release: Question pairs.
\newblock \emph{data. quora. com}, 2017.

\bibitem[Kairouz et~al.(2021)Kairouz, Liu, and Steinke]{kairouz2021distributed}
Peter Kairouz, Ziyu Liu, and Thomas Steinke.
\newblock The distributed discrete gaussian mechanism for federated learning with secure aggregation.
\newblock In \emph{International conference on machine learning}, pages 5201--5212. PMLR, 2021.

\bibitem[Li et~al.(2025)Li, Yang, Wu, Wang, Ohtsuki, Fu, Pan, and Shen]{li2025mobillm}
Liang Li, Xingke Yang, Wen Wu, Hao Wang, Tomoaki Ohtsuki, Xin Fu, Miao Pan, and Xuemin Shen.
\newblock Mobillm: Enabling llm fine-tuning on the mobile device via server assisted side tuning.
\newblock \emph{arXiv preprint arXiv:2502.20421}, 2025.

\bibitem[Lin et~al.(2024)Lin, Hu, Zhang, Chen, Fang, Chen, Li, Vepakomma, and Gao]{lin2024splitlora}
Zheng Lin, Xuanjie Hu, Yuxin Zhang, Zhe Chen, Zihan Fang, Xianhao Chen, Ang Li, Praneeth Vepakomma, and Yue Gao.
\newblock Splitlora: A split parameter-efficient fine-tuning framework for large language models.
\newblock \emph{arXiv preprint arXiv:2407.00952}, 2024.

\bibitem[McMahan et~al.(2017)McMahan, Moore, Ramage, Hampson, and y~Arcas]{mcmahan2017communication}
Brendan McMahan, Eider Moore, Daniel Ramage, Seth Hampson, and Blaise~Aguera y~Arcas.
\newblock Communication-efficient learning of deep networks from decentralized data.
\newblock In \emph{Artificial intelligence and statistics}, pages 1273--1282. PMLR, 2017.

\bibitem[Rajpurkar et~al.(2016)Rajpurkar, Zhang, Lopyrev, and Liang]{rajpurkar2016squad}
Pranav Rajpurkar, Jian Zhang, Konstantin Lopyrev, and Percy Liang.
\newblock Squad: 100,000+ questions for machine comprehension of text.
\newblock In \emph{Proceedings of the 2016 conference on empirical methods in natural language processing}, pages 2383--2392, 2016.

\bibitem[Ren et~al.(2024)Ren, Wei, Xia, Su, Cheng, Wang, Yin, and Huang]{ren2024representation}
Xubin Ren, Wei Wei, Lianghao Xia, Lixin Su, Suqi Cheng, Junfeng Wang, Dawei Yin, and Chao Huang.
\newblock Representation learning with large language models for recommendation.
\newblock In \emph{Proceedings of the ACM Web Conference 2024}, pages 3464--3475, 2024.

\bibitem[Socher et~al.(2013)Socher, Perelygin, Wu, Chuang, Manning, Ng, and Potts]{socher2013recursive}
Richard Socher, Alex Perelygin, Jean Wu, Jason Chuang, Christopher~D Manning, Andrew~Y Ng, and Christopher Potts.
\newblock Recursive deep models for semantic compositionality over a sentiment treebank.
\newblock In \emph{Proceedings of the 2013 conference on empirical methods in natural language processing}, pages 1631--1642, 2013.

\bibitem[Stevens et~al.(2022)Stevens, Skalka, Vincent, Ring, Clark, and Near]{stevens2022efficient}
Timothy Stevens, Christian Skalka, Christelle Vincent, John Ring, Samuel Clark, and Joseph Near.
\newblock Efficient differentially private secure aggregation for federated learning via hardness of learning with errors.
\newblock In \emph{31st USENIX security symposium (USENIX Security 22)}, pages 1379--1395, 2022.

\bibitem[Su et~al.(2024)Su, Prakash, Chen, Gong, Yu, Fu, and Pan]{su2024dafl}
Huai-an Su, Pavana Prakash, Rui Chen, Yanmin Gong, Rong Yu, Xin Fu, and Miao Pan.
\newblock Dafl: Device-to-device transmissions for delay efficient federated learning over mobile devices.
\newblock \emph{IEEE Internet of Things Journal}, 2024.

\bibitem[Sun et~al.(2024)Sun, Li, Li, and Ding]{sun2024improving}
Youbang Sun, Zitao Li, Yaliang Li, and Bolin Ding.
\newblock Improving lora in privacy-preserving federated learning.
\newblock \emph{arXiv preprint arXiv:2403.12313}, 2024.

\bibitem[Sung et~al.(2022)Sung, Cho, and Bansal]{sung2022lst}
Yi-Lin Sung, Jaemin Cho, and Mohit Bansal.
\newblock Lst: Ladder side-tuning for parameter and memory efficient transfer learning.
\newblock In \emph{Proc. of Advances in Neural Information Processing Systems (NeurIPS)}, New Orleans, Louisiana, December 2022.

\bibitem[Tian et~al.(2022)Tian, Wan, Lyu, Yao, Jin, and Sun]{tian2022fedbert}
Yuanyishu Tian, Yao Wan, Lingjuan Lyu, Dezhong Yao, Hai Jin, and Lichao Sun.
\newblock Fedbert: When federated learning meets pre-training.
\newblock \emph{ACM Transactions on Intelligent Systems and Technology (TIST)}, 13\penalty0 (4):\penalty0 1--26, 2022.

\bibitem[Vepakomma et~al.(2020)Vepakomma, Singh, Gupta, and Raskar]{vepakomma2020nopeek}
Praneeth Vepakomma, Abhishek Singh, Otkrist Gupta, and Ramesh Raskar.
\newblock Nopeek: Information leakage reduction to share activations in distributed deep learning.
\newblock In \emph{2020 International Conference on Data Mining Workshops (ICDMW)}, pages 933--942. IEEE, 2020.

\bibitem[Wang et~al.(2018)Wang, Singh, Michael, Hill, Levy, and Bowman]{wang2018glue}
Alex Wang, Amanpreet Singh, Julian Michael, Felix Hill, Omer Levy, and Samuel~R Bowman.
\newblock Glue: a multi-task benchmark and analysis platform for natural language understanding. corr abs/1804.07461 (2018).
\newblock \emph{arXiv preprint arXiv:1804.07461}, 2018.

\bibitem[Wang et~al.(2024)Wang, Shen, He, Sun, Wang, Lyu, and Li]{wang2024flora}
Ziyao Wang, Zheyu Shen, Yexiao He, Guoheng Sun, Hongyi Wang, Lingjuan Lyu, and Ang Li.
\newblock Flora: Federated fine-tuning large language models with heterogeneous low-rank adaptations.
\newblock \emph{Advances in Neural Information Processing Systems}, 37:\penalty0 22513--22533, 2024.

\bibitem[Warstadt et~al.(2019)Warstadt, Singh, and Bowman]{warstadt2019neural}
Alex Warstadt, Amanpreet Singh, and Samuel~R Bowman.
\newblock Neural network acceptability judgments.
\newblock \emph{Transactions of the Association for Computational Linguistics}, 7:\penalty0 625--641, 2019.

\bibitem[Williams et~al.(2017)Williams, Nangia, and Bowman]{williams2017broad}
Adina Williams, Nikita Nangia, and Samuel~R Bowman.
\newblock A broad-coverage challenge corpus for sentence understanding through inference.
\newblock \emph{arXiv preprint arXiv:1704.05426}, 2017.

\bibitem[Wolf et~al.(2019)Wolf, Debut, Sanh, Chaumond, Delangue, Moi, Cistac, Rault, Louf, Funtowicz, et~al.]{wolf2019huggingface}
Thomas Wolf, Lysandre Debut, Victor Sanh, Julien Chaumond, Clement Delangue, Anthony Moi, Pierric Cistac, Tim Rault, R{\'e}mi Louf, Morgan Funtowicz, et~al.
\newblock Huggingface's transformers: State-of-the-art natural language processing.
\newblock \emph{arXiv preprint arXiv:1910.03771}, 2019.

\bibitem[Xu et~al.(2024)Xu, Cai, Wu, Li, and Wang]{xu2024fwdllm}
Mengwei Xu, Dongqi Cai, Yaozong Wu, Xiang Li, and Shangguang Wang.
\newblock $\{$FwdLLM$\}$: Efficient federated finetuning of large language models with perturbed inferences.
\newblock In \emph{2024 USENIX Annual Technical Conference (USENIX ATC 24)}, pages 579--596, 2024.

\bibitem[Yang et~al.(2025)Yang, Li, Wan, Li, Qi, Liu, Ohtsuki, Fu, and Pan]{yang2025pae}
Xingke Yang, Liang Li, Zhiyi Wan, Sicong Li, Xiaoqi Qi, Jiang Liu, Tomoaki Ohtsuki, Xin Fu, and Miao Pan.
\newblock Pae mobillm: Privacy-aware and efficient llm fine-tuning on the mobile device via additive side-tuning.
\newblock \emph{arXiv preprint arXiv:2507.01216}, 2025.

\bibitem[Ye(2024)]{ye2024cross}
Qinyuan Ye.
\newblock Cross-task generalization abilities of large language models.
\newblock In \emph{Proceedings of the 2024 Conference of the North American Chapter of the Association for Computational Linguistics: Human Language Technologies (Volume 4: Student Research Workshop)}, pages 255--262, 2024.

\bibitem[Zaken et~al.(2021)Zaken, Ravfogel, and Goldberg]{zaken2021bitfit}
Elad~Ben Zaken, Shauli Ravfogel, and Yoav Goldberg.
\newblock Bitfit: Simple parameter-efficient fine-tuning for transformer-based masked language-models.
\newblock \emph{arXiv preprint arXiv:2106.10199}, 2021.

\bibitem[Zhang et~al.(2022)Zhang, Roller, Goyal, Artetxe, Chen, Chen, Dewan, Diab, Li, Lin, et~al.]{zhang2022opt}
Susan Zhang, Stephen Roller, Naman Goyal, Mikel Artetxe, Moya Chen, Shuohui Chen, Christopher Dewan, Mona Diab, Xian Li, Xi~Victoria Lin, et~al.
\newblock Opt: Open pre-trained transformer language models.
\newblock \emph{arXiv preprint arXiv:2205.01068}, 2022.

\bibitem[Zhang et~al.(2023)Zhang, Yang, Dai, Wang, Yu, Qu, and Xu]{zhang2023fedpetuning}
Zhuo Zhang, Yuanhang Yang, Yong Dai, Qifan Wang, Yue Yu, Lizhen Qu, and Zenglin Xu.
\newblock Fedpetuning: When federated learning meets the parameter-efficient tuning methods of pre-trained language models.
\newblock In \emph{Annual Meeting of the Association of Computational Linguistics 2023}, pages 9963--9977. Association for Computational Linguistics (ACL), 2023.

\bibitem[Zhu et~al.(2019)Zhu, Liu, and Han]{zhu2019deep}
Ligeng Zhu, Zhijian Liu, and Song Han.
\newblock Deep leakage from gradients.
\newblock \emph{Advances in neural information processing systems}, 32, 2019.

\end{thebibliography}
\bibliographystyle{plainnat}

\newpage
\appendix

\section{Further Experimental Results and Discussion}\label{app:Discussion}

\subsection{Server Storage} 

\begin{table}[!htbp]
\centering
\small
\caption{Server cache size across backbones and tasks (homogeneous Xavier clusters).}
\label{tab:server_cache_size}
\setlength{\tabcolsep}{10pt}
\begin{tabular}{lccc}
\toprule
\textbf{Backbone} & \multicolumn{3}{c}{\textbf{Server cache size (GB)}} \\
\cmidrule(lr){2-4}
 & \textbf{RTE} & \textbf{SST2} & \textbf{MNLI} \\
\midrule
OPT-1.3B     & 0.23 & 6.21 & 36.77  \\
RoBERTa-Base & 0.08 & 2.40 & 14.21 \\
\bottomrule
\end{tabular}
\end{table}

The major focus of device-server collaborative LLM fine-tuning works is to reduce computation, memory, and/or communication costs on resource-constrained mobile devices. Most state-of-the-art studies~\citep{li2025mobillm, yang2025pae, xu2024fwdllm} assume that the server has sufficient resources. We follow a similar setting and prioritize the analysis of task performance and system efficiency from the mobile-device side in the main paper.

Here, we provide quantitative measurements of EC-MobiLLM’s server-side cache usage across different backbone sizes (from RoBERTa-Base to OPT-1.3B) and dataset scales (from RTE to MNLI). As shown in Table~\ref{tab:server_cache_size}, under the 1.3B model setting, the server-side cache is about 36.77 GB for the largest dataset MNLI (about 393K examples), while it is only about 0.23 GB for a small-scale dataset such as RTE (about 2K examples). Since each device uploads only a small amount of selected activations, the storage overhead introduced by server-side caching does not make EC-MobiLLM impractical, supporting its deployability. Given current mobile memory constraints and the fact that per-user datasets are usually relatively limited, these results indicate that, in our target scenario, the server-side storage cost is acceptable and does not raise serious feasibility concerns for the current design.

However, when targeting larger backbones or higher data throughput, edge servers with tight storage budgets may require additional operational mechanisms, such as cache lifecycle management, eviction policies, and compression. We view these as meaningful directions for future work.

\subsection{Impact of Network Speed on Training Efficiency} 

\begin{figure}[!htbp]
	\centering
	\includegraphics[width=0.50\linewidth]{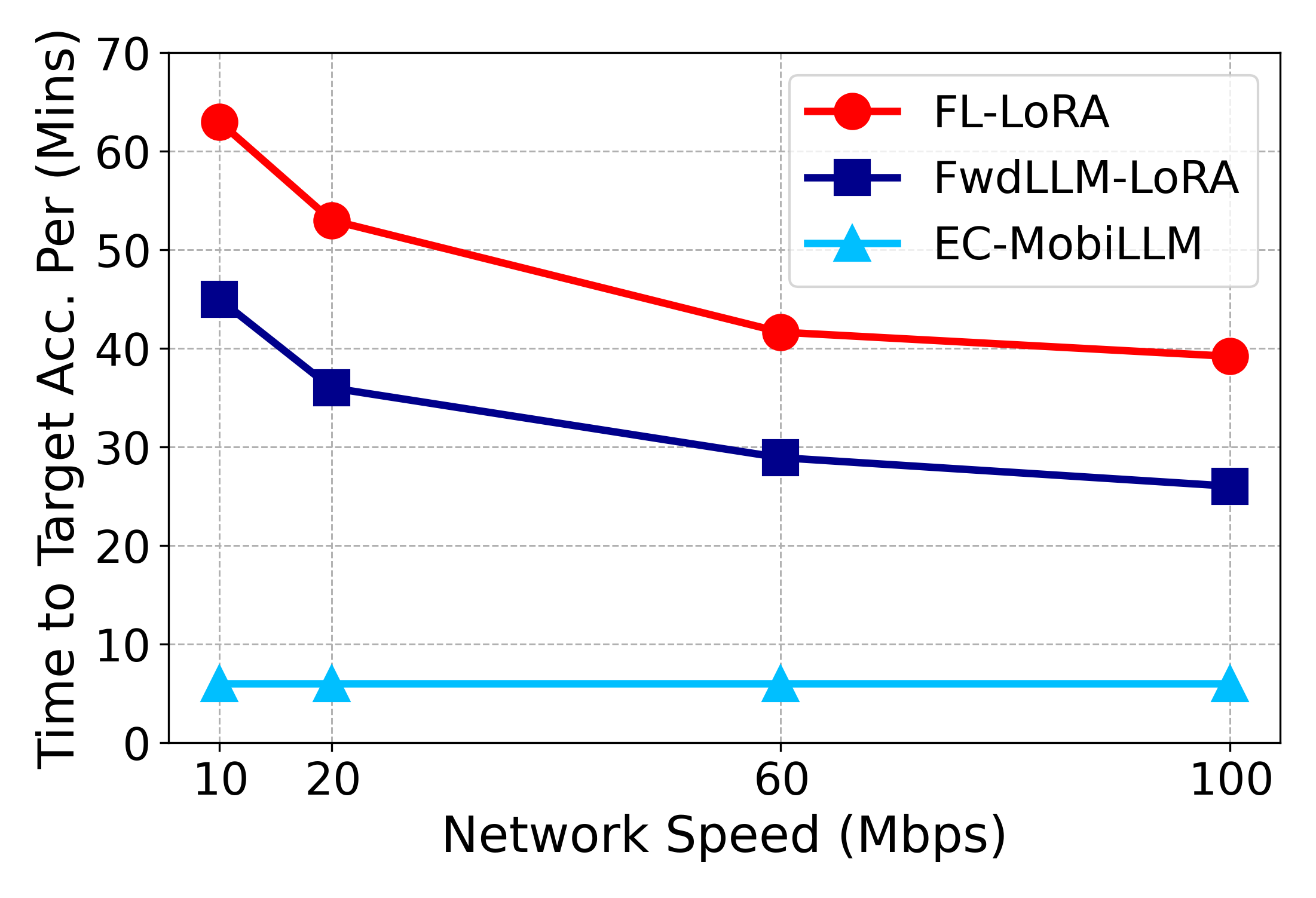}
        \caption{Impact of network speed on time-to-accuracy (homogeneous Xavier clusters, task: RoBERTa-Base@MRPC).}
        \label{fig:time-networkspeed}
\end{figure}

In practical collaborative LLM fine-tuning, device-server communication makes training efficiency sensitive to network conditions. Consequently, network fluctuations may impact overall training efficiency. Figure~\ref{fig:time-networkspeed} illustrates the training efficiency across methods under different wireless transmission speeds. Since the total training time in FL-LoRA and FwdLLM-LoRA is affected by the communication time, they suffer severe performance degradation under low-speed transmission conditions. For example, FwdLLM-LoRA shows a $1.62\times$ slowdown at 10Mbps vs. 100Mbps. In contrast, EC-MobiLLM maintains stable total latency across different transmission speeds, demonstrating almost no variations from 10 Mbps to 100 Mbps. Such resilience stems from parallel scheduling that overlaps communication with device- and server-side computation, masking transmission delays with on-device compute time that is unaffected by network-speed fluctuations.


\subsection{Privacy Discussion on Activation Transmission}\label{app:Privacy}

In this paper, the main contribution of EC-MobiLLM is to provide a collaborative LLM fine-tuning framework for heterogeneous mobile devices. It focuses on reducing on-device computation, memory, and communication cost, improving training efficiency, and enabling cross-backbone collaboration across devices. EC-MobiLLM involves the transmission of intermediate activations during training. This is aligned with the privacy assumptions commonly adopted in split learning related architectures~\citep{tian2022fedbert} and federated learning (FL)~\citep{mcmahan2017communication} systems, where devices transmit gradients/activations instead of raw user text data, thereby avoiding the direct exposure of original user data.

However, when data security is strictly considered, prior studies~\citep{zhu2019deep,geiping2020inverting,vepakomma2020nopeek} have shown that gradients or activations, as representations derived from local data, may still be used to infer original input information. To mitigate such risks, existing works~\citep{kairouz2021distributed,stevens2022efficient} have proposed additional protection mechanisms, such as differential privacy (DP) mechanisms~\citep{dwork2006differential} and secure multiparty computation (SMC)~\citep{du2001secure}. DP mechanisms can be applied by adding calibrated Gaussian or Laplace noise to uploaded gradients, activations, outputs, or objective functions, thereby reducing the risk of privacy leakage from transmitted information. SMC can be used as a complementary cryptographic mechanism to further protect the communication and computation process. These protection mechanisms can also be applied to EC-MobiLLM’s activation transmission.

Compared with conventional FL/SFL-style methods, EC-MobiLLM has a favorable property from the perspective of privacy-loss composition. Conventional methods usually require multiple communication rounds, where repeated releases of noisy gradients or activations lead to accumulated privacy loss. In contrast, EC-MobiLLM uploads the selected activations for each local sample only once, and the server subsequently reuses the cached activations for iterative training. This reduces repeated data-dependent releases and can mitigate privacy-loss accumulation. Under the same overall privacy budget, one-time uploading can therefore reduce the required noise level and better preserve fine-tuning performance.

To evaluate this property, we conduct a sample-level DP experiment on SST-2 under the heterogeneous-device setting. We compare EC-MobiLLM without privacy protection, one-shot caching that uploads privatized activations once and reuses them for 20 training epochs, and a repeated-release variant that independently perturbs and uploads activations in every epoch. Both private configurations use the same overall DP budget $(\epsilon,\delta)=(4,10^{-5})$.

\begin{table}[!htbp]
\centering
\small
\setlength{\tabcolsep}{5pt}
\caption{Fine-tuning performance under differential privacy on SST-2 across heterogeneous mobile devices. Accuracy is reported in percentage (\%) $\uparrow$.}
\label{tab:dp_evaluation}
\begin{tabular}{lccccc}
\hline
Method
& \shortstack{Releases\\per Sample}
& DP Budget
& \shortstack{AGX\\(OPT-1.3B)}
& \shortstack{Xavier\\(OPT-350M)}
& \shortstack{TX2\\(OPT-125M)} \\
\hline
EC MobiLLM (Non-private)       & -- & --                  & 92.9 & 92.4 & 92.1 \\
EC MobiLLM (One-shot Cache)    & 1  & $(4,10^{-5})$       & 89.2 & 88.5 & 88.1 \\
Repeated Releases Variants & 20 & $(4,10^{-5})$       & 86.5 & 85.3 & 84.4 \\
\hline
\end{tabular}
\end{table}

As shown in Table~\ref{tab:dp_evaluation}, introducing DP protection leads to an average accuracy reduction of about 3.9 percentage points for EC-MobiLLM with one-shot caching compared to the non-private setting. However, under the same overall DP budget, one-shot caching consistently outperforms repeated releases by 2.7--3.7 percentage points across the three devices. These results demonstrate that EC-MobiLLM can incorporate DP protection while reducing the impact of privacy-preserving noise on fine-tuning performance through one-time activation uploading and server-side cache reuse.

\subsection{Ablation Study of Key Components}
\label{app:component_ablation}

To evaluate the contribution of each key component in EC-MobiLLM, we conduct an ablation study under the same experimental setting as Table~2. Specifically, we separately remove token selection (TS), server caching (SC), and non-blocking asynchronous collaboration (NBAC) to examine their effects on system efficiency and fine-tuning performance.

\begin{table}[!htbp]
\centering
\small
\setlength{\tabcolsep}{4pt}
\caption{Ablation study of the key components in EC-MobiLLM (RoBERTa-Base@MRPC, heterogeneous devices). TTA is measured at 86.5\% accuracy.}
\label{tab:component_ablation}
\begin{tabular}{lccccc}
\hline
Method
& \shortstack{On-device\\Memory (GB)$\downarrow$}
& \shortstack{On-device\\Comp. (TFLOPs)$\downarrow$}
& \shortstack{On-device\\Comm. (MB)$\downarrow$}
& \shortstack{FT Acc.\\(\%)$\uparrow$}
& \shortstack{TTA\\(min)$\downarrow$} \\
\hline
EC-MobiLLM & 0.38 & 2.7 & 1.1 & 88.1 & 5.9 \\
w/o. TS    & 0.38 & 2.7 & 281.6 & 88.1 & 45.7 \\
w/o. SC    & 0.38 & 54.0 & 22.0 & 88.1 & 91.3 \\
w/o. NBAC  & 0.38 & 2.7 & 1.1 & 88.1 & 9.6 \\
\hline
\end{tabular}
\end{table}

As shown in Table~\ref{tab:component_ablation}, removing TS significantly increases communication cost and training time due to the transmission of full activations. Without SC, devices need to repeat forward propagation and activation transmission across training epochs, resulting in 20$\times$ higher computation and communication costs. Removing NBAC increases TTA from 5.9 to 9.6 minutes due to synchronization delays. Meanwhile, all configurations maintain the same on-device memory usage and fine-tuning accuracy. These results demonstrate that the three components jointly improve the communication, computation, and training efficiency of EC-MobiLLM without sacrificing fine-tuning performance.

\subsection{Inference-Time Overhead}
\label{app:inference_overhead}

To evaluate the inference overhead introduced by EC-MobiLLM, we compare backbone-only inference with inference using the additional projection modules and shared side-network on three heterogeneous mobile devices. We measure the inference latency, peak on-device memory, and computation cost under both configurations.

\begin{table}[!htbp]
\centering
\small
\setlength{\tabcolsep}{4pt}
\caption{Inference-time overhead of EC-MobiLLM on heterogeneous mobile devices. ``+ Modules'' denotes adding the projection modules and shared side-network to the backbone. The relative increases are shown in parentheses.}
\label{tab:inference_overhead}
\begin{tabular}{llccc}
\hline
Device / Backbone
& Configuration
& \shortstack{Latency\\(s/batch)$\downarrow$}
& \shortstack{Peak Memory\\(GB)$\downarrow$}
& \shortstack{Comp.\\(TFLOPs/batch)$\downarrow$} \\
\hline
TX2 / OPT-125M
& Backbone Only & 0.98 & 0.56 & 0.36 \\
& Backbone + Modules & 1.03 (+5.1\%) & 0.57 (+1.8\%) & 0.37 (+2.8\%) \\
\hline
Xavier NX / OPT-350M
& Backbone Only & 1.07 & 1.10 & 1.29 \\
& Backbone + Modules & 1.12 (+4.7\%) & 1.12 (+1.8\%) & 1.32 (+2.3\%) \\
\hline
AGX Xavier / OPT-1.3B
& Backbone Only & 1.89 & 3.44 & 5.05 \\
& Backbone + Modules & 1.95 (+3.2\%) & 3.48 (+1.2\%) & 5.20 (+3.0\%) \\
\hline
\end{tabular}
\end{table}

As shown in Table~\ref{tab:inference_overhead}, the additional projection modules and side-network increase inference latency by only 3.2\%--5.1\%, peak on-device memory by 1.2\%--1.8\%, and computation cost by 2.3\%--3.0\% across the three devices. These results demonstrate that EC-MobiLLM introduces limited inference overhead while supporting collaborative fine-tuning across heterogeneous backbones.

\section{Datasets Statistics} \label{app:dataset}

\begin{table}[!htbp]
\centering
\caption{Datasets Statistics.}
\label{tab:benchmarks}
\small
\setlength{\tabcolsep}{8pt}
\begin{tabular}{l l l l}
\hline
\textbf{Dataset} & \textbf{Description} & \textbf{Task} & \textbf{\# Samples (train/validation)} \\
\hline
CoLA  & Linguistic Acceptability & Classification & 8551 / 1043 \\
SST2  & Sentiment Analysis & Classification & 67349 / 872 \\
MRPC  & Sentence Equivalence & Classification & 3668 / 408 \\
STSB  & Sentence Similarity & Regression & 5749 / 1500 \\
QQP   & Paraphrase Recognition & Classification & 363846 / 40430 \\
RTE   & Textual Entailment & Classification & 2490 / 277 \\
QNLI  & Natural Language Inference & Classification & 104743 / 5463 \\
MNLI  & Textual Entailment & Classification & 392702 / 9815 (9832) \\
DialogSum & Abstract Summary & Generation & 12460 / 500 \\
\hline
\end{tabular}
\end{table}

\section{Computation of Reported Results} \label{app:calc}

In this section, we use Table~\ref{table:method_comparison} as an example to present how we compute \emph{on-device resource overhead} and \emph{training efficiency} metrics for EC-MobiLLM and other baselines. We consider a setting with 100 homogeneous NVIDIA Xavier clients, each of which loads the same RoBERTa-Base backbone. The unified experimental setup is as follows: Dataset is evenly distributed across 100 clients according to the standard federated learning configuration. FP16 precision, batch size = 8, sequence length = 256, and an in-lab Wi-Fi throughput of 60 Mbps.

\subsection{On-Device Memory footprint} 

\begin{itemize}
\item \textbf{FL-LoRA:} Performs full LoRA fine-tuning on the RoBERTa model on the device. The memory footprint consists of model weights + intermediate activation + optimizer states and hence is the largest among all methods.

\item \textbf{SFL-LoRA:} To keep raw data and labels on the device, only offload the middle 10 layers of the 12-layer transformer to the server; the remaining layers are LoRA fine-tuned on the device. The memory footprint consists of partial weights + partial intermediate activation + partial optimizer states, thus lower than FL-LoRA.

\item \textbf{FwdLLM-LoRA and EC-MobiLLM:} The device only performs forward propagation. The memory footprint consists of model weights + a small inference-time intermediate activation, achieving the smallest memory footprint.

\end{itemize}

\subsection{On-Device computation} 

We report on-device computation as the total FLOPs executed on a single device over the entire training process until it reaches the target accuracy. Let the number of global rounds be \(R\), the number of local iterations (epochs/passes) per round be \(E\), and let \(\mathrm{Cost}_{\text{local}}\) denote the computation for one complete local epoch under a given method. 

\begin{itemize}
\item \textbf{FL-LoRA:} Each local epoch performs the full RoBERTa + LoRA (forward \& backward).
\[
\mathrm{Computation}_{\text{device}} \;=\; \mathrm{Cost}_{\text{FT(full\ backbone,\ LoRA)}} \times R \times E.
\]

\item \textbf{SFL-LoRA:} Only the non-offloaded layers performs LoRA on device (forward \& backward).
\[
\mathrm{Computation}_{\text{device}} \;=\; \mathrm{Cost}_{\text{FT(partial\ backbone,\ LoRA)}} \times R \times E.
\]

\item \textbf{FwdLLM-LoRA (forward-only with perturbations):} The device performs only forward propagation; each round uses \(K\) local perturbation forwards.
\[
\mathrm{Computation}_{\text{device}} \;=\; \mathrm{Cost}_{\text{Infer(full\ backbone)}} \times R \times K.
\]

\item \textbf{EC-MobiLLM:} The device performs a single forward propagation traversal of its local data (no local replay).
\[
\mathrm{Computation}_{\text{device}} \;=\; \mathrm{Cost}_{\text{Infer(full\ backbone)}} \
.\]

\end{itemize}

\subsection{On-Device Communication}

We report on-device communication as the total amount of data a single device exchanges with the server over the entire training process until it reaches the target accuracy. Let the number of global rounds be \(R\), the number of local iterations (epochs/passes) per round be \(E\).

\begin{itemize}
\item \textbf{FL-LoRA:}
In standard FL, devices and server exchange trainable parameters in both directions (upload/downlink) each global round:
\[
\mathrm{Communication}_{\text{device}}
= 2 \times R \times \bigl|\theta_{\text{LoRA}}\bigr|,
\]
where \(\bigl|\theta_{\text{LoRA}}\bigr|\) is the size of the on-device LoRA trainable parameters.

\item \textbf{SFL-LoRA:}
Beyond the round-wise parameter exchange, split learning requires frequent exchange of intermediate activations and backward gradients during local forward/backward:
\[
\mathrm{Communication}_{\text{device}}
= 2 \times R \times \bigl|\theta_{\text{LoRA}}\bigr|
\;+\; R \times E \times \mathrm{Comm}_{\text{act/grad-per\ epoch}},
\]
where \(\mathrm{Comm}_{\text{act/grad-per\ epoch}}\) denotes the activation/gradient traffic for one local forward+backward epoch.

\item \textbf{FwdLLM-LoRA.}
Similar to FL-LoRA, the standard communication rhythm for federated learning:
\[
\mathrm{Communication}_{\text{device}}
= 2 \times R \times \bigl|\theta_{\text{LoRA}}\bigr|.
\]

\item \textbf{EC-MobiLLM:}
The device performs a single forward propagation traversal of its local data and uploads only a small subset of layer-wise activations selected by the token selector; there is no parameter round-trip or gradient return:
\[
\mathrm{Communication}_{\text{device}}
\;\approx\; \mathrm{Comm}_{\text{selected-activations- single\ epoch}}.
\]
\end{itemize}

\subsection{Training Time to Target Accuracy}

We report the training time to target accuracy for each LLM federated FT framework. Let the number of global rounds be \(R\), and define:
\(t_1\) — per-round on-device runtime,
\(t_2\) — per-round device–server parameter communication time,
\(t_3\) — per-round server aggregation time.

For \textbf{FL-LoRA / SFL-LoRA / FwdLLM-LoRA}, the per-round steps run \emph{serially}, so
\[
T_{\text{train}} = (t_1 + t_2 + t_3)\times R.
\]
The only difference lies in $t_1$:
\begin{itemize}
\item \textbf{FL-LoRA:} $t_1=$ on-device LoRA fine-tuning (forward + backward).
\item \textbf{SFL-LoRA:} $t_1=$ split-learning based device--server co-training; frequent activation/gradient exchange makes $t_1$ typically longer.
\item \textbf{FwdLLM-LoRA:} $t_1=$ on device multiple forward propagations.
\end{itemize}

\noindent\textbf{EC-MobiLLM} does not follow a federated sequential rhythm. Instead, it overlaps single forward propagation time on the device side with communication time and server iteration time in parallel. The total training time approximates the maximum duration among these three components. In our testing, the server iteration typically dominates, i.e., $T_{\text{train}}\approx t_{\text{server-iter}}$.





\end{document}